%% file: acl_latex.tex
\documentclass[11pt]{article}

\usepackage[final]{acl}

\usepackage{times}
\usepackage{latexsym}

\usepackage[T1]{fontenc}

\usepackage[utf8]{inputenc}

\usepackage{microtype}

\usepackage{inconsolata}

\usepackage{graphicx}

\usepackage{xspace}
\usepackage{xurl}
\usepackage{booktabs}
\usepackage{amsmath}
\usepackage{soul}

\usepackage{subcaption} 

\input{commands}

\newcommand{\repo}{\url{https://github.com/g8a9/ouvia}\xspace}
\newcommand{\ethicsinstitution}{Instituto Superior Técnico\xspace}

\newcommand{\acro}{\textsc{Ouvia}\xspace}

\newcommand{\uswhite}{US\textsubscript{W}\xspace}
\newcommand{\usblack}{US\textsubscript{B}\xspace}
\newcommand{\hindi}{Hindi\xspace}

\title{\acro: A User-centered Framework for Measuring Usability of \\ Speech Translation in Real-World Communication Scenarios}

\usepackage{upgreek}
\newcommand{\fbk}{$^{\Upphi}$}          %
\newcommand{\itc}{$^{\Uplambda}$}         %
\newcommand{\ist}{$^{\Upsigma}$}        %
\newcommand{\cmu}{$^{\Uppi}$}          %
\newcommand{\umd}{$^{\Upupsilon}$}      %
\newcommand{\tp}{$^{T}$}           %

\author{Giuseppe Attanasio\itc, Beatrice Savoldi\fbk, Daniel Chechelnitsky\cmu \\
 \vspace{0.2cm}
\textbf{Matteo Negri\fbk, Marine Carpuat\umd, Maarten Sap\cmu, André F.T. Martins\ist\itc\tp} \\ 
 \itc~Instituto de Telecomunicações, 
 \fbk~Fondazione Bruno Kessler, \cmu~Carnegie Mellon University \\
 \umd~University of Maryland, \ist~Instituto Superior Técnico, Universidade de Lisboa, \tp~TransPerfect \\
 \url{gattanasio.work@gmail.com}
}

\begin{document}
\maketitle
\begin{abstract}

Speech translation (ST) is increasingly adopted in user applications, yet its evaluation largely focuses on decontextualized testbeds and holistic quality, rather than end users' communication needs.
We introduce \acro, an evaluation framework for measuring user-perceived \textit{usability} of speech translation outputs in real-world settings.
\acro focuses on one-to-one communication: an English speaker needs to convey a request to a Portuguese speaker, and the message is automatically translated.
Through a custom web app and multi-phase study design, we collect more than $1{,}750$ such interactions in healthcare and everyday situations, mediated by four ST systems, involving speakers from three English dialects and two genders.
We find that modern ST serves people only to a limited extent---only around half of interactions are rated as usable---with significant gaps in reported usability across demographic groups.
Moreover, among quality metrics, we find that QA-based evaluation is a substantially stronger predictor of real-world usability than standard approaches.
Together, these findings stress the importance of situated, user-centered evaluation frameworks that go beyond holistic quality scores and attend to who the technology serves---and how well.\footnote{Study platform and data under CC-BY 4.0 at \repo.}
\end{abstract}

\section{Introduction}

Speech translation now underpins a broad range of
user-facing applications, from cross-lingual real-time conferencing to software for multilingual education\footnote{\url{https://www.wordly.ai/education-translation}} or clinical healthcare triage.\footnote{\url{https://www.caretotranslate.com/}}
A key catalyst of this progress is the field's ability to \textit{evaluate} translation systems rigorously, whether through targeted evaluation campaigns \cite{iwslt-ws-2025-1} or scalable automatic quality metrics \cite[e.g.,][]{rei-etal-2022-comet,han-etal-2024-speechqe,juraska-etal-2025-metricx}.

While laudable, these efforts address only a fraction of what evaluation demands.
Campaigns typically run \textit{in vitro}: they present decontextualized (non-situated) test segments to participants and treat quality as a holistic concept---detached from function and purpose \cite{colina2009further,liu-etal-2024-evaluation} and primarily serving system performance tracking and ranking rather than actual real-world use.
Moreover, automatic quality assessments tell little about ramifications---e.g., whether bad translations imply clinical risk \cite{khoong2019assessing}, or quality gaps reflect real-world allocative harms \cite{savoldi-etal-2024-harm}---and risks amplifying social biases \cite{zaranis-etal-2025-watching}.
Purely automatic, decontextualized, holistic evaluation cannot be the sole end. It should, instead, be supported by new forms of inquiries, centered around end-users---
today, largely laypeople \cite{nurminen2021investigating}---their communication needs, and real-world uses and situations \cite{savoldi-etal-2025-translation,carpuat-etal-2025-interdisciplinary}.

\begin{figure*}[!t]
    \centering
    \includegraphics[width=0.8\linewidth]{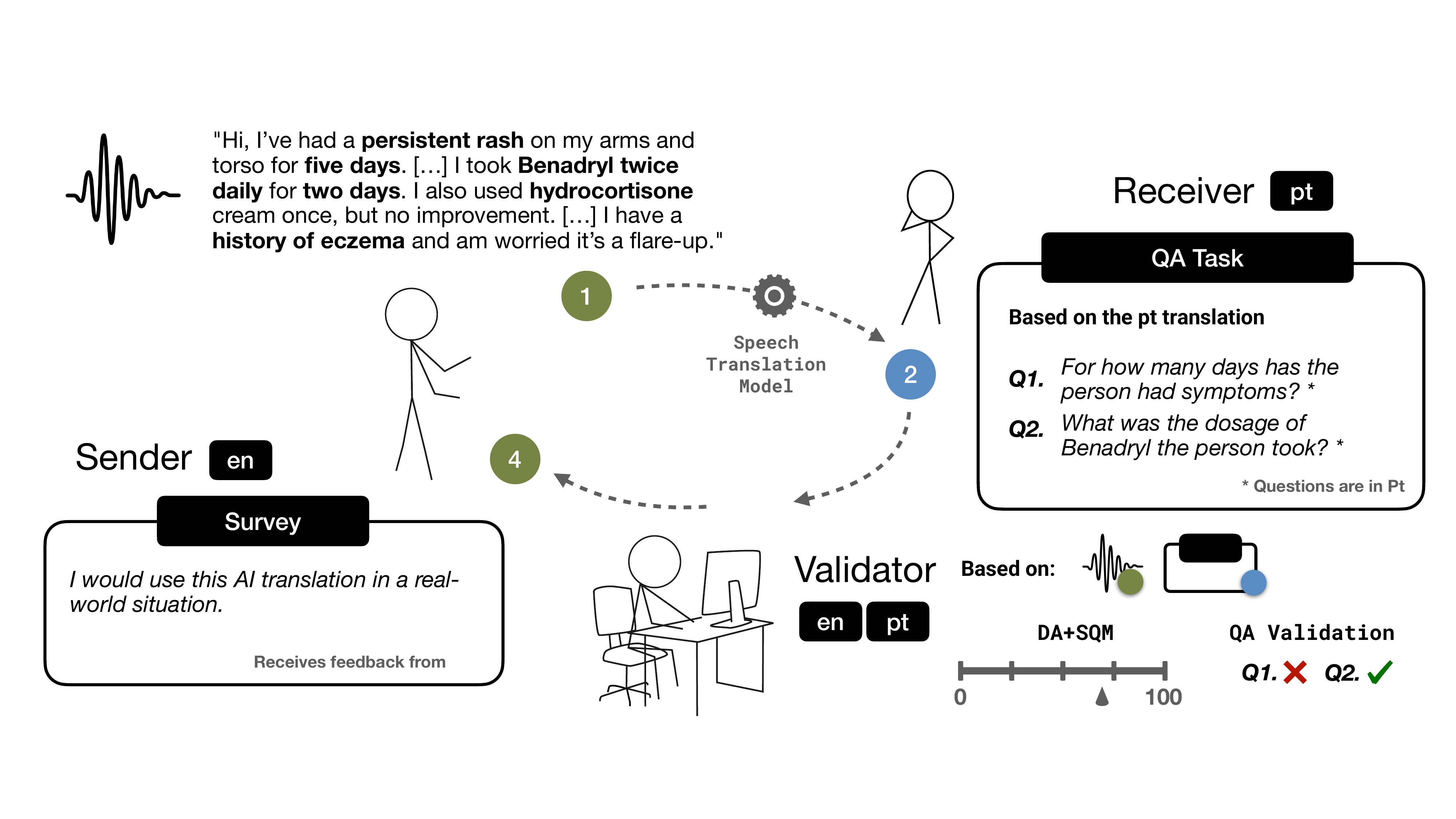}
    \caption{\textbf{Study Workflow.} \senderchip{1}: A English-speaking \textit{sender} reads aloud and records a conversation starter, which we translate with a speech translation system. \receiverchip{2}: A L1 Portuguese \textit{receiver} answers up to 10 open-text questions about the translated text. \validatorchip{3}: A validator, fluent in both languages, assesses translation quality and flags the questions answered incorrectly in \receiverchip{2}. \senderchip{4}: the initial sender rates their perceived \textit{usability} compounding satisfaction, trust, and reliance in two time points: first based on the translation from \senderchip{1}, then with the additional outcomes from \validatorchip{3}.}
    \label{fig:study-overview}
\end{figure*}

This paper introduces \acro,\footnote{Past tense of Portuguese \textit{ouvir}, to hear.} an evaluation framework for capturing user-perceived \textit{usability} of speech translation systems in real-world communication scenarios.
We pursue three overarching goals:
(1) capture lay users' perceptions of the usability of translation outputs across different high- and low-stakes situations; 
(2) measure whether speaker dialect and gender identity give rise to gaps in translation quality and usability, and how these relate;
(3) investigate the predictive power of (automatic) quality metrics for usability, and characterize the practical implications of metric choice.
\acro bundles the following contributions:

\textbf{A conceptual evaluation framework} that moves ST evaluation from decontextualized testbeds to situated, user-centered assessment. We devise a multi-phase protocol that mimics ST-mediated one-to-one interactions (see Figure~\ref{fig:study-overview}): a \textit{sender} initiates a conversation with an English request, which is automatically translated for a Portuguese \textit{receiver} who answers comprehension questions based on the translation. A third bilingual \textit{validator} annotates the exchange with quality signals. Finally, the sender is surveyed on several dimensions of usability. This design captures what in-vitro evaluation cannot: whether a translation is \textit{usable} for a specific person, in a specific situation, to accomplish a specific communicative goal.

\textbf{A user study, data, and findings.} We implement the framework through a user study grounded in English-to-Portuguese translation in Portugal---a highly multicultural country where large segments of the immigrant population rely on English as a \textit{lingua franca} \cite{aima2025}. We recruit native English speakers across different dialects \citep{doi:10.1073/pnas.1915768117} and Hindi speakers as a non-native group---the largest migrant population in Portugal after Brazilians and the fastest-growing, for whom mediated translation is particularly relevant. We collect $N=1{,}738$ subjective judgments from $174$ unique speakers on translations from four state-of-the-art ST systems across healthcare and everyday scenarios. Our findings show that \textbf{current systems serve real-world users only to a limited extent}, that \textbf{usability gaps exist across speaker demographics}, and that \textbf{QA-based evaluation predicts usability substantially better than standard coarse-grained metrics}---reinforcing the need for situated, user-centered evaluation of ST.

\textbf{A custom web platform and speech corpus.} To run the study, we develop a web platform supporting speech recordings and  multi-user interactions. To support future research, we release $14.6$ hours of prompted speech recordings, $13.8$K+ human annotations for QA-based quality evaluation, and $1.7$K+ translation quality scores.

\section{Study Design}

We focus on \textit{communication} scenarios \cite{hovy2002principles}: a speaker wants to correctly convey a message to someone in a different language.
In this setting, correctness might not necessarily mean generic translation quality. Rather, the exchange might be successful ``as long as people get the information they want, [\ldots] or manage to convey their intentions'' \cite{hutchins2005current}. 
To study this setup at scale, we approximate such an interaction through an online study with a novel multi-phase design. 
For every interaction, we recruit two participants who role-play a verbal exchange\footnote{We acknowledge the distinction between role-playing and real-world users \citep{ai-etal-2007-comparing}. While we prioritize a controlled, online setup, future study may fill this gap with in-the-wild data collection.} and a third-party bilingual validator to independently assess the correctness and quality of the translation mediating the exchange.

\subsection{Workflow}
\label{ssec:study-workflow}

Each study unit consists of four sequential phases.

\textbf{Phase \senderchip{1}.} A \textit{sender} records a passage we collected (\S\ref{ssec:compiling-starters}). This \textit{conversation starter} is 40-60 words long and contains key information, such as named entities or quantities, that a system should translate correctly (see bold phrases in Figure~\ref{fig:study-overview}). 

\textbf{Phase \receiverchip{2}.} We translate the 
sender's recording automatically and show the outcome to a \textit{receiver}, asking them to reply to a set of open-ended questions grounded in the original passage. (\S\ref{ssec:compiling-questions}).

\textbf{Phase \validatorchip{3}.} We recruit a bilingual \textit{validator} to verify the translation in two ways. \textit{(i)} A direct assessment score \cite{graham-etal-2013-continuous}---Direct Assessment with Scalar Quality Metrics \cite[DA+SQM]{kocmi-etal-2022-findings}---where we provide 
a slider from 0 to 100, marked with seven labeled tick marks indicating different quality labels that capture accuracy and grammatical correctness (details in Appendix~\ref{sec:app:SQM}). \textit{(ii)} An assessment for each question indicating if the receiver answered it correctly.

\textbf{Phase \senderchip{4}.} 
We give the automatic translation to the initial sender and task them with a survey to collect their subjective judgments about the translation.
Specifically, they rate their agreement on a 5-point scale (1 = Strongly Disagree, 5 = Strongly Agree) with statements aimed to assess: \textit{i) satisfaction} perceived with the translation output (adapted from \citet{basoah2025not}); \textit{ii) trust}, the belief that the output will be beneficial (adapted from \citet{hoffman-trust}); and \textit{iii) reliance}, the self-reported behavioral intention to depend on the output in practice (adapted from \citet{McGrath-trust-reliance}).\footnote{The exact statements are: \textit{i)} ``I am satisfied with the quality of the AI translation''; \textit{ii)} ``I trust the AI translation to convey my message''; \textit{iii)}: ``I would use this AI translation in a real-world situation''.}
Crucially, we encourage them to answer as if they were in a real-life situation needing to 
communicate with someone who speaks Portuguese. Together, these judgments provide a situated indication of three dimensions of a translation \textit{\textbf{usability}} (\S\ref{sec:analysis_setup}).\footnote{Although usability is traditionally defined in HCI as the effectiveness, efficiency, and satisfaction with which users achieve goals with a specific system \citep{iso9241-11-2018}, here we use it more broadly to capture how useful a user finds a translation in a given communicative context.}

Since senders do not speak Portuguese, we administer the survey at two time points. 
First, we capture their baseline---their unconditioned tendency toward the translation usability based solely on the source passage, their recording, and the Portuguese translation they cannot linguistically verify. Upon completion, we disclose the outcomes of Phases \receiverchip{2} and \validatorchip{3}, enabling senders to make another, but informed, final judgment.

\subsection{Compiling Starters}
\label{ssec:compiling-starters}

For a realistic condition in \senderchip{1}, we aim for conversation starters that are truthful in content and style to real-life interactions, are of high quality,
and cover both high- and low-stakes scenarios. Moreover, we prioritize information-dense starters, i.e., passages that include multiple key information. To increase diversity, starter data are sampled from existing corpora as well as automatically generated. We confirm their quality via manual validation, as described in Appendix \ref{app:validation}.
\footnote{See Appendix~\ref{ssec:app:starters-preparation} for full details on automatic data cleaning and generation of both high- and low-stakes sets.}

\textbf{Healthcare Scenarios.}
We target the healthcare domain, where automatic translation finds frequent use \cite{genovese2024artificial} and errors have profound ramifications \cite{taira2021pragmatic}. 
We sample entries from MED-MT \cite{fareez2022dataset}. 
The dataset contains transcripts of simulated patient-physician interviews, primarily focusing on respiratory cases.
Since the original transcripts contain filled pauses and disfluencies, we use Gemini 2.5 Pro to strip them and improve fluency without changing the core content. To increase content and style diversity, 
prompted Qwen 3 32B and Gemini Pro 2.5 to create over-the-counter pharmacy conversation starters, which mentions to symptoms, their duration, medication names, and allergies.

\textbf{Everyday Scenarios.}
We source customer requests from BConTrasT \cite{farajian-etal-2020-findings}, which consists of customer support dialogues.  
As in the healthcare case, we 
collect synthetic LLM-generated starters expressing
requests about 
everyday scenarios, such as booking a taxi, banking requests, or hotel check-in.

Overall, we sample 50 items each from both MT-MED and BConTrasT and generate 100 samples per domain (50 from each LLM), collecting a total of 300 unique conversation starters. Appendix Table~\ref{tab:starters-examples} reports examples and statistics.

\subsection{Compiling Questions}
\label{ssec:compiling-questions}

Largely inspired by translation evaluation via question answering \citep{ki-etal-2025-askqe}, we automatically generate up to 10 unique questions for each starter. 
 Following \citet{fernandes2025llms}, who show improved question relevance when generating in English, we generate questions in English before translating them into Portuguese. For both generation and translation steps, we use Gemini 2.5 Pro,  prompting the model to ground the question in the starter's main information (see an example in Figure~\ref{fig:study-overview}, \receiverchip{2}). See Appendix Figure~\ref{fig:prompt_questions} and \ref{tab:question_translation} for the prompt used.
We validate quality, as reported in Appendix~\ref{app:validation}. Out of 200 questions, only one was answerable. Errors are thus extremely rare, but we account for them in the validation stage of our study: receivers can flag questions as unanswerable and validators can accept this as correct.

\subsection{Speech Translation Systems}

We experiment with four translation systems based on open-weight models that are currently state of the art \cite{papi2025hearing}.
We include three direct speech translation models, Phi 4 Multimodal \cite{abouelenin2025phi}, Voxtral Small \cite{liu2025voxtral}, and DesTA2 \cite{lu2025developing}, and one cascaded system that uses Whisper large-v3 \cite{radford2023robust} to transcribe the recordings followed by Tower+ 9B \cite{rei2025tower+}, a strong language model specialized for translation in European languages.
After a sender completes \senderchip{1}, we pick randomly one translation system using default inference parameters and prompts to avoid introducing confounders. 
See generation details in Appendix~\ref{app:ssec:translation-details}.

\subsection{Participant Recruitment}
\label{ssec:participant-recruitment}

We recruited study participants on Prolific.\footnote{\url{https://www.prolific.com/}. See Appendix~\ref{app:sec:participant-recruitment} for details on recruitment.}. We require the first language of \textit{receivers} to be Portuguese and proficiency in both English and Portuguese for \textit{validators}. 
To explore the impact of native English dialects and non-native accented speech, we recruited \textit{senders} from three language groups based on (self-declared) nationality and ethnic group:\footnote{We acknowledge ``ethnicity'' is a fraught concept, but use it to match the screening filters available on Prolific.} (1) US White and (2) US Black speakers, whose first language is English, and (3) Hindi native speakers, who are proficient in English as a second language.
Within each group, we balanced participants by gender identity.
Using the platform's filters, we select ``Man'' and ``Woman.''\footnote{
We discuss the motivations and implications of this binary setup in \textbf{Ethical Considerations.}
}

Since 21\% of the senders dropped out of the study (Phases \senderchip{1} and \senderchip{4} were, on average, six days apart), we recruited replacement participants only to complete the survey in \senderchip{4}. We saw no significant difference between third- and first-person survey scores (see Appendix~\ref{app:ssec:first-third-person} for details).

\subsubsection{Participant-Task Assignment}
\label{ssec:task-assignments}

For senders, we use a between-subjects design with respect to group membership \citep{charness2012experimental}: each participant belongs to exactly one language-by-gender combination, with 30 participants recruited per cell. 
Cell sizes were determined to support adequately powered comparisons \citep{brysbaert2019many}.
The stimuli (starters) are shared across all groups, enabling both within- and across-group comparisons. Each participant is assigned 10 starters, stratified across domain (healthcare, mundane), source type (synthetic, natural), and model. Stratified assignment preserves internal validity by ensuring no individual participant is disproportionately exposed to any domain or model, thus preventing individual biases.
For receivers, we recruit 20 participants per language group and randomly assign them 30 translations each. Similarly, we recruit 20 participants who each validate 30 exchanges.
To prevent practice effects, we randomize the order of items displayed to each participant.

\input{tab_scores}
\input{fig_usability_drilldown}

\section{Analysis Setup}
\label{sec:analysis_setup}

\paragraph{Cohorts.} Our study results can be analyzed across several cohorts, identified by three sender language groups (US White: \uswhite, US Black: \usblack, \hindi speakers), two genders (Woman, Man), conversation topic (Health, Everyday), source (Natural, Synthetic), and translation model.
To formalize the relationships between these factors and their impact on usability, we describe them in a causal graph: the sender, starter's source and topic, and translation model all contribute to the translation quality, etc.\footnote{We acknowledge the existence of other contingent factors---e.g., the age of participants, the recording device, background noise---which we do not control in this study.} The graph is shown in Appendix Figure~\ref{fig:causal-graph}.

\paragraph{Quality Control on Recordings.} Prompted speech collected through online platforms is prone to recording artifacts due to (device or environment) noise, microphone activation delays, or unfamiliarity with recording software that can cause the first milliseconds of an utterance to be clipped \citep{lee-etal-2025-speech}. We controlled for these issues with a semi-automatic procedure. We first transcribed all recordings with Whisper Large v3 and flagged entries meeting either of two conditions: the first or last three words (after normalization) do not match the prompt, or WER exceeds $0.6$. One author then manually reviewed all flagged entries to eliminate false positives, and entries that failed verification were discarded. We excluded participants with more than three faulty entries entirely. These checks led to the exclusion of two women and two men from \uswhite, one woman and one man from \usblack plus two individual records from \usblack, and two women and three men from \hindi. 
The final corpus counts $N=1{,}738$ observations across $174$ senders, $60$ receivers, and $69$ validators.

\paragraph{Usability ($u$).} Prior literature establishes that trust and satisfaction can predict reliance~\cite{McGrath-trust-reliance, bhattacherjee2001understanding}, which led us to expect that senders would rate these three dimensions similarly within a given translation. To verify this, we run a standard factor analysis, confirming that more than 90\% of the variance in the scores is explained by a single underlying factor (see Appendix~\ref{app:ssec:factor-analysis}).
For practical purposes, we average the three scores at the instance level, creating a single compound usability score $u$ based on the informed final judgments (\S\ref{ssec:study-workflow}), which we use as the target variable of interest in the study. We compare it to the \textit{baseline} $u$ in an upcoming analysis.  

\paragraph{Quality Metrics.} When studying the correlation between translation quality and usability, we rely on the following set of automatic metrics, established and widely used in the NLP literature for their high agreement with human assessment of quality: COMET \cite{rei-etal-2020-comet}, COMET Kiwi Base and XL \cite{rei-etal-2022-comet}, XCOMET-XL \cite{guerreiro-etal-2024-xcomet}, and MetricX 24 XL \cite{juraska-etal-2024-metricx}.
We also include the judgments from \validatorchip{3}: Translation Score and QA Score, defined as the ratio of correct answers given in \receiverchip{2}.

\section{Findings}

We begin by investigating our main interest: whether our participants have found the translations \textit{usable} (\S\ref{ssec:findings-usability}). Upon noting differences across groups, we analyze the effect of demographic factors (\S\ref{ssec:demographic-effects}) and translation quality (\S\ref{ssec:translation-quality}). Finally, we study how different ``quality'' signals relate to user-perceived usability (\S\ref{ssec:which-quality-signal}).

\input{fig_validator-translation-score}

\subsection{Do lay users find ST outputs useful?}
\label{ssec:findings-usability}

\textbf{Not fully}, with most usability scores falling between 3 and 4 in a 5-point Likert scale. Table~\ref{tab:survey-scores} reports descriptive statistics across language groups, including the (final) usability $u$ scores compared to the \textit{baseline} scores. Figure~\ref{fig:usability-score-distribution-group} shows the $u$ score distributions. 

\uswhite score usability the highest in both the baseline ($u_b=4.06$) and final conditions ($u=3.87$) while \hindi the lowest ($3.76$ and $3.35$, respectively).
When comparing baseline and final assessments, all groups decrease their scores, and the distribution of their scores also increases in variance. Therefore, the quality signals mitigate prior assumptions and increase the polarization of judgment.   
This result is expected, as this signal likely surfaces errors more evidently to them.
However, such an effect varies across language groups. \hindi lower their scores ($-0.40$) more than twice as much as \uswhite do ($-0.19$), an effect likely mediated by receiving poorer quality translations (discussed in \S\ref{ssec:demographic-effects}).
The two factors, hence, compound: \uswhite and \hindi start from uneven baselines, and this gap is exacerbated when becoming aware of external quality signals.
We do not record the same phenomenon between \uswhite and \usblack.

\subsection{How do demographics and contextual factors impact usability?}
\label{ssec:demographic-effects}

\textbf{Both contextual factors and demographics affect usability significantly.} We fit a linear mixed regression model (LMM) to predict usability ($u$) using the translation model, starter topic, source, and baseline $u$ as fixed effects, and sender, validator, and receiver as random effects. Figure~\ref{fig:usability-coeff} shows some of the resulting coefficients (full details in Appendix~\ref{app:ssec:lmm}). The largest variance is explained by the translation model, with Voxtral and Tower+ yielding the best scores ($p<0.05$; see per-model $\mu(u)$ in Appendix Figure~\ref{fig:usability-scores-translation-system}).
Scores on Health-related starters are significantly higher than Everyday (+$0.33$, $p<0.05$). 

Interestingly, the LMM discloses that language variant-based differences (Table~\ref{tab:survey-scores}) are not significant. However, we observe a negative effect of Gender: Woman ($-0.2$, $p<0.1$) and a positive intersectional effect for the intersection of \uswhite x Woman ($+0.36$ over the reference \hindi x Man, $p<0.05$). 
As this finding suggests that within-language group differences might exist, we plot the marginal means estimated by the LMM in Figure~\ref{fig:usability-interaction-group-gender}. The figure offers one additional core insight: gaps in usability across groups are dampened by similar results on men; women, instead, scored translation unevenly. The biggest difference is between \uswhite and \hindi, where usability is $4.18$ and $3.69$, respectively---a significant gap ($p<0.05$ if computing pairwise differences, after Holm correction).
Together, these findings suggest that a combination of factors---led by the choice of the translation model and the intersectional demographic group---determines self-perceived usability. 
Moreover, these usability gaps across demographic groups raise a natural question:

\subsection{Are starters from certain senders systematically mistranslated?}
\label{ssec:translation-quality}

To address this question, we focus on the validator's SQM quality scores---one of the two human judgments of translation quality our study provides (more in \S\ref{ssec:which-quality-signal}).
We fit a new LMM to predict it and inspect the model coefficients (see Figure~\ref{fig:validator-score}).
Compared to \hindi, validators score \uswhite and \usblack higher by $13.71$ ($p<0.05$) and $7.43$ ($p<0.1$) points, respectively. Marginal means (Figure~\ref{fig:vs-group-gender}) highlight these gaps better, where \uswhite men score $76.4$, against \hindi men's $62.7$. 
Interestingly, within \hindi---the group with the worst quality according to validators---women score higher than men, despite later expressing lower usability scores (Figure~\ref{fig:usability-interaction-group-gender}).
Moreover, Figure~\ref{fig:vs-model} reports the means by model and topic. As expected, translation quality also varies significantly by translation model, with Voxtral and Tower+ leading the board and Health receiving higher scores overall. 

Yet validator SQM scores are just one lens, and an expensive one. The question is whether cheaper, automated signals tell the same story.

\subsection{Do translation quality signals predict usability?}
\label{ssec:which-quality-signal}

\begin{table}[!t]
\small
\centering
\setlength{\tabcolsep}{4pt}
\begin{tabular}{@{}p{3.2cm}rrrr@{}}
\toprule
\textbf{Metric} & \multicolumn{1}{l}{\textbf{e}} & \multicolumn{1}{l}{\textbf{e\textsubscript{low}}} & \multicolumn{1}{l}{\textbf{e\textsubscript{med}}} & \multicolumn{1}{l}{\textbf{e\textsubscript{high}}} \\ \midrule
MetricX 24 & 2.35 & 1.35 & 0.74 & 1.03 \\
XCOMET XL & 1.94 & 1.13 & 0.79 & 0.82 \\
COMET & 2.85 & 1.27 & *0.58 & 0.65 \\
COMET Kiwi & 2.43 & 1.37 & *0.47 & 1.35 \\
COMET Kiwi XL & 2.69 & 1.44 & 0.77 & 1.33 \\ \addlinespace[0.5em]
Translation Score & 2.11 & 0.69 & 0.55 & 0.32 \\
QA Score & \textbf{2.94} & \textbf{1.61} & \textbf{2.64} & \textbf{3.06} \\ \bottomrule
\end{tabular}
\caption{\textbf{Quality metrics effect on usability ($u$).} From a LMM (\textbf{e} column) and from OLS models fitted within low, medium, and high translation score tertiles. An effect of $k$ means that moving from the worst- to the best-observed quality on that metric is associated with a $k$-point shift in usability (1--5 scale), holding other covariates constant. All metrics rescaled to $[0,1]$. All effects but (*) are significant ($p<0.01$).}
\label{tab:effects-metrics-regimes}
\end{table}

\textbf{Partially, and QA is superior to coarse-grained overall judgments}.
We compute simple Spearman rank correlation against $u$ 
and note a variable degree across metrics (see Appendix Figure~\ref{fig:metrics-correlation}). Among automatic metrics, XCOMET XL is the highest ($\rho=0.49$), and COMET is the lowest ($\rho=0.42$), whereas Human judgments have reasonably higher correlation.  
However, QA Score has $\rho=0.63$, which is interestingly higher than translation Score ($\rho=0.56$).

To further investigate these differences, we conduct two more measurements. (1) We fit one LMM per metric using the same fixed- and random-effects setup as in \S\ref{ssec:findings-usability} adding the metric to covariates, and storing the resulting coefficient---this number measures the metric's effect on usability. (2) We split the records following the Translation Score's tertiles, identifying three quality regimes; for each, we fit one Ordinary Least Squares (OLS) regression on $u$ (more robust than LMM for setups with fewer data points), again rotating the metrics and recording the coefficient. These numbers tell us how closely each metric tracks usability at varying quality.  
Table~\ref{tab:effects-metrics-regimes} reports these coefficients. 
We find that automatic metric effects decline as quality increases. Translation Score shows a similar trend, whereas QA confirms the stronger effect, outscoring all metrics across all regimes. This finding suggests that fine-grained content-based evaluation tracks usability more consistently than a single overall quality score at varying quality levels.

\section{Discussion}

Our results call for renewed interest in human-centered evaluation of ST. We demonstrate that user-perceived usability and coarse-grained translation quality are related but distinct constructs---and that this distinction matters. 
This is where \acro adds value: by situating evaluation in a real communicative exchange, it surfaces nuances that aggregate quality scores obscure---from which systems and demographic groups benefit, to where standard metrics fail as deployment signals.

\paragraph{The choice of ST is consequential, and newer is not always better.} For realistic use cases, direct ST is not always the right solution: both Phi 4 and DeSTA2, despite being recently released, lag behind a simpler cascade pipeline that combines a smaller foundation speech recognition model (Whisper v3) with a larger, highly specialized translation model (Tower+) (see Figure~\ref{fig:vs-model} for details). 
This finding underscores that ``new'' does not guarantee ``usable,'' and that practitioners should evaluate systems on ecologically valid tasks rather than standard benchmarks alone.

\paragraph{Demographic gaps persist and have real stakes.} We find significant usability gaps along both language variety and gender. This finding aligns with prior work documenting inequitable ST performance on standard benchmarks \cite{fucci-etal-2025-different}, but our results extend this concern to systems in actual deployment scenarios. As shown in Figure~\ref{fig:usability-drilldown}a, more than $66\%$ of \uswhite speakers find the system usable ($u\geq4$), compared to only $49\%$ for \usblack speakers and $43\%$ for \hindi speakers. These are not abstract disparities: because our usability measure explicitly captures willingness to rely on the translation in real-world situations, what is at stake is whether a speaker will use ST technology for consequential communication, depending on the variety of English they speak.

\paragraph{QA-based evaluation tracks usability better than standard quality metrics.} Translation quality predicts usability, but not all quality signals are equally informative. Standard quality estimation metrics rank highly on leaderboards for overall quality prediction~\cite{kocmi-etal-2025-findings}, yet in communicative settings, they explain only a fraction of variation in user-perceived usability. QA score, 
which measures whether key information, such as named entities and quantities, is translated correctly, is a substantially better predictor. Crucially, it is also more robust:
in the high-quality regime, differences in standard automatic metrics poorly distinguish between translations that users find more or less usable, whereas differences in QA score do. Figure~\ref{tab:effects-metrics-regimes} illustrates this contrast directly. 

\begin{figure}[!t]
    \centering
    \includegraphics[width=1\linewidth]{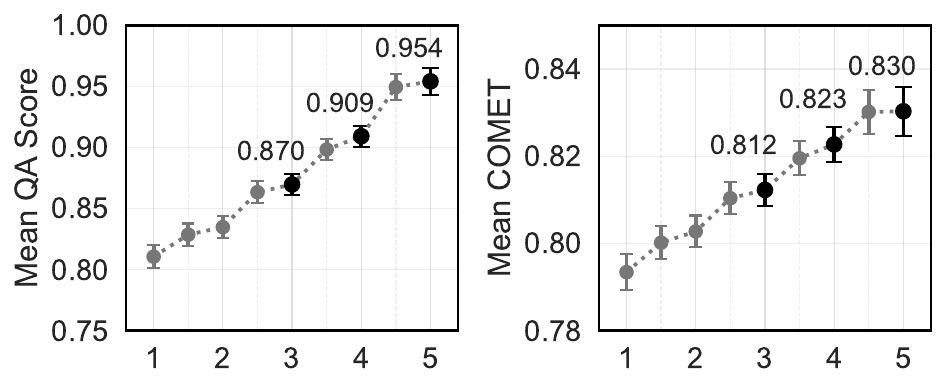}
    \caption{\textbf{Mean QA Score and COMET ($\pm95\%$ CI)} among records whose usability meets or exceeds each threshold value on the x-axis. Error bars are 95\% confidence intervals. Annotations at thresholds 3, 4, and 5.}
    \label{fig:usability-informed-qa-comet}
\end{figure}

\paragraph{Toward usability-grounded deployment thresholds.} Translation quality metrics are not designed to capture usability, so their limited predictive power is unsurprising. They also remain opaque: providing a generic overall score with mostly comparative power---model X outperforms model Y---but offering little sense of whether, and how effectively, a translation serves a user's communicative needs \citep{savoldi-etal-2025-translation}. Yet 
usability human assessment is not scalable. We thus leverage our study to anchor existing metrics to a concrete usability threshold. 
In Figure~\ref{fig:usability-informed-qa-comet}, we map QA Score---a potentially automatable evaluation signal---and COMET---the metric with the highest impact on usability (Table~\ref{tab:effects-metrics-regimes})---onto the usability scores they correspond to. At $u$ = $4$, a reasonable threshold for a translation a lay user would satisfactorily rely on in practice,  a QA Score of $\sim$$0.91$ and a COMET score of $\sim$$0.82$ can serve as practical deployment benchmarks. Notably, unlike COMET, QA score remains more meaningful beyond this point, thus adding further evidence in favor of its discriminative power. When human judgments are unavailable, these thresholds can offer a principled, usability-grounded basis for system selection and deployment decisions.

\section{Related Work}

Our work sits at the intersection of user-centered approaches to machine translation (MT) \citep{carpuat-etal-2025-interdisciplinary} and demographic disparities in speech technologies \citep{gaido-etal-2020-breeding, savoldi-etal-2022-dynamics, savoldi-etal-2022-morphosyntactic, zhang2022mitigating, harris-etal-2024-modeling, attanasio-etal-2024-twists}. For MT, prior work has explored user needs in realistic settings across various domains, including medical \citep{mehandru-etal-2023-physician}, migration \citep{valdez2025google}, housing, and employment \citep{robertson2022understanding}.
\citet{xiao-etal-2025-toward} shows that users overly trust MT outputs with mistranslations, and \citet{ki-etal-2025-share} focuses on quality feedback that helps users calibrate MT reliance in communication settings. \citet{savoldi-etal-2024-harm} further demonstrates that automatic metrics fail to capture the real-world human and economic costs of gender bias in MT. 
Most recently, studying goal achievement through effective communication in MT-mediated regimes, \citet{mohammed2026far} investigate the impact of discourse-related errors and the extent to which automatic quality metrics capture them. 

Calls for grounding evaluation in user needs and societal relevance are growing \citep{liebling-etal-2022-opportunities, savoldi-etal-2025-translation, ungless-etal-2025-amplifying}, yet remain only partially addressed in ST---the emerging frontier for human-to-human mediated communication. 
Early studies of supportive ST tools exist. \citet{belvin-etal-2005-transonics} proposed a speech-to-speech system for realistic medical dialogues. \citet{janakiram2020user} measured user satisfaction with a custom speech-enabled fixed-phrase translator in a hospital emergency context.  
Our work aligns with these interests and brings these concerns together, revealing the limits of ST for communication across accents and those of coarse-grained assessments for real-world usability. Our findings point toward QA-based assessment \citep{ki-etal-2025-askqe, fernandes2025llms}---which targets accurate transmission of key information rather than closeness to a human reference---as a stronger proxy for the real-world usability that lay users need in situated, communicative scenarios.

\section{Conclusion}

We have introduced Ouvia, a new evaluation framework to study the user-perceived usability of several state-of-the-art ST systems. We described how we designed a multi-phase user study that underpins our research questions, collecting $N=1,738$ judgments from $174$ speakers across diverse demographic groups. Our results show that current systems serve real-world users only to a limited extent, that usability gaps persist across speaker demographics, and that QA-based evaluation is a substantially stronger predictor of usability than standard automatic metrics. Together, these findings reinforce the need for human-centered, situated evaluation of ST---one that goes beyond holistic quality scores and attends to who the technology serves, in what context, and to what end.

\section*{Limitations} \label{sec:limitations}

\paragraph{Language pair.} Reliable assessments require sizable data samples and participant pools, which we prioritize during budget allocation. For this reason, we limit our study to one language pair (English - European Portuguese), while including speakers across genders and language variants motivated by realistic usage scenarios of immigrants in Portugal. We remain cautious about generalizing our findings to language pairs that are underserved by current speech technologies, where usability gaps may be more pronounced.

\paragraph{Scripted and clean speech.}
Our conversation starters are grammatically fluent, and senders are instructed to read them aloud without fillers, repetitions, or disfluencies. While this allows us to control for content quality, it departs from naturalistic spoken communication, where such features could affect downstream translation. 
Similarly, we ask senders to record in noise-free environments, resulting in clean, read speech that differs from the acoustic conditions typical of real-world use---such as background noise. Thus, these conditions likely favor current speech translation systems, meaning our usability estimates may be optimistic relative to deployment scenarios.
While---to the best of our knowledge---our study is the first to approximate a realistic communication setting for evaluating speech technologies, future work should examine usability in more spontaneous settings.

\paragraph{Crowdworkers.} Our validators are bilingual crowdworkers rather than professional translators. Still, non-expert can provide reliable assessment using standardized scales \citep{kocmi-etal-2022-findings}. Furthermore, our binary QA validation task requires comprehension rather than translation expertise.

\paragraph{Usability and QA.}
QA-based evaluation is a strong predictor of usability in our data, particularly in high-quality regimes, pointing to its potential as a future evaluation proxy. Still, further work is needed to understand this relationship across other communicative settings and with fully automated pipelines. Notably, our QA assessment is manual and disclosed to the sender---yet so is the validator's holistic quality score, which correlates substantially less with usability. This suggests it is the fine-grained, information-targeted nature of QA that drives its predictive power, not the human involvement or disclosure alone. 
Moreover, QA-based evaluation in \acro is simplified by (\textit{i}) the existence of ground truth conversation starters, which we use to extract the questions, but that might not be generally available, and (\textit{ii}) the starters' content itself, which is, by design, information-rich in our study.

\section*{Ethical Considerations} \label{sec:ethics}

\paragraph{Intended Use and Participation.} \acro is designed for evaluating the usability of ST systems across diverse demographic groups. All participants were recruited through Prolific and financially compensated according to the platform's standards. The study was approved by the Ethics Committee of \ethicsinstitution.
All participants provided informed consent prior to participation.

\paragraph{Voice Privacy.} 
Voice data carries inherent re-identification risks: a speaker may be recognized even absent explicit metadata, and recordings can in principle be used for voice cloning. Our study design limits, but does not eliminate, these risks: we collect a small number of short segments per participant (10 segments, each 30s or less), which is insufficient for training high-fidelity speaker-specific synthesis models, though we note that contemporary zero-shot cloning methods can operate on only a few seconds of reference audio. Participants gave informed consent to public release. The dataset is distributed under CC BY 4.0 through a gated request form, where requesters provide an email address and acknowledge that the data should not be used for speaker identification or voice cloning. As CC BY permits redistribution, this acknowledgment functions as a statement of intended use rather than an enforceable restriction.

\paragraph{Gender.} Gender is among the most salient perceptual traits of one's identity \citep{fuchs2010differences, zimman2021gender}, and gendered differences represent a central concern in sociophonetic research \citep{zimman2020sociophonetics} and speech technologies \citep{attanasio-etal-2024-twists, sanchez2024beyond}. Following Prolific's screening filters, we operationalize gender as a self-declared attribute using \textit{Man} and \textit{Woman} as categories, each subsuming both cisgender and transgender speakers. We initially included a \textit{Non-Binary} option (as available in Prolific) for our two largest cohorts (US White and US Black). However, we ultimately excluded it: the category is an overly broad umbrella term that conflates distinct gender identities, making group-level comparisons theoretically unsound. This was further reflected in low uptake (four participants). Recruiting through queer and allied communities to improve participation and gender diversity beyond the binary remains an important direction for future work.

\section*{Acknowledgments}

We thank Sweta Agrawal, whose support was instrumental in devising and implementing the work; the European Association for Machine Translation for sponsoring the user study and data collection; and the members of the SARDINE Lab (Lisbon) for piloting early versions of the annotation platform.
Giuseppe Attanasio and André Martins were supported by the AMALIA project under Measure RE-C05-i08 of the Portuguese national Programa de Recuperação e Resiliência, the Portuguese Recovery and Resilience Plan through project C645008882-00000055 (Center for Responsible AI), the project DECOLLAGE (ERC-2022-CoG 101088763), FCT/MECI through national funds, and cofunded EU funds under UID/50008: Instituto de Telecomunicações.
Beatrice Savoldi has received funding from the European Union’s Horizon research and innovation programme under grant agreement No. 101135798, project Meetween (My Personal AI Mediator for Virtual MEETings BetWEEN People). 
Matteo Negri has received funding from the European Union’s Horizon Europe programme under grant agreement No. 101213369 (project DVPS).

\bibliography{minimal}

\appendix

\section{Participant Recruitment}
\label{app:sec:participant-recruitment}

We recruited all participants through \url{https://www.prolific.com/}, an established platform to recruit crowdworkers. To join the study, each participant had to read and accept the Terms and Conditions, which had been approved in advance by an Ethics Committee. 
Participants could join the study only once and in a single role. All participants were paid 9 GBP/h.
All participants were allowed to join the study using a smartphone, tablet, or notebook.
The following is a list of criteria to join the study:

\begin{itemize}
    \item \textbf{\uswhite Sender}: Nationality: USA, First Language: English, Gender: Man/Woman (including transgender), Broad Ethnic Group: White, Approval Rate: $75$--$100$.
    \item \textbf{\usblack Sender}: Nationality: USA, First Language: English, Gender: Man/Woman (including transgender), Broad Ethnic Group: Black/Afroamerican, Approval Rate: $75$--$100$.
    \item \textbf{\hindi Sender}: First Language: Hindi, Fluent Language: English, Gender: Man/Woman (including transgender), Approval Rate: $75$--$100$.
    \item \textbf{Receiver}: First Language: Portuguese, Gender: Man/Woman (including transgender), Approval Rate: $75$--$100$.
    \item \textbf{Validator}: First Language: Portuguese, Fluent Language: English, Gender: Man/Woman (including transgender), Approval Rate: $75$--$100$.
\end{itemize}

\subsection{First- vs. Third-Perspective Assessments}
\label{app:ssec:first-third-person}

To validate the use of replacement participants in cases where original senders dropped out of the longitudinal study, we examined whether subjective usability scores differed depending on whether the assessment was given from a first-person perspective (senders evaluating their own recorded interactions, $N=1{,}358$) or a third-person perspective (replacement participants evaluating interactions they had not produced, $N=380$). We fitted a linear mixed-effects model (REML) predicting the average survey score from perspective alongside translation quality, comprehension rate, translation model, gender, and language group as fixed effects, with random intercepts for participant and conversation starter. The perspective coefficient was negligible and non-significant ($\beta = 0.009$, $p = .873$), indicating no meaningful difference in subjective ratings between the two perspectives. These results justify the replacement strategy: crowdworkers assessing conversations they did not themselves produce provided ratings statistically indistinguishable from those of the original senders.

\begin{figure*}[!t]
    \centering
    \includegraphics[width=0.9\linewidth]{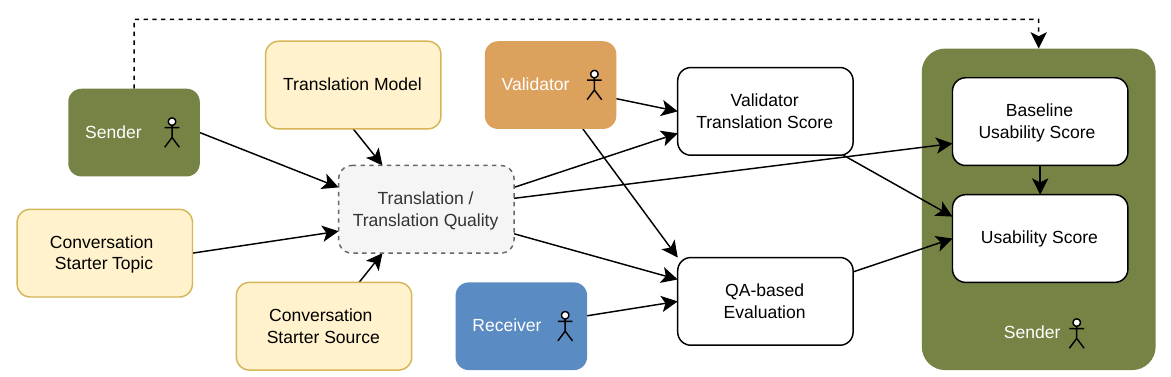}
    \caption{\textbf{Causal Graph.} The links establish how different factors interact in the study. Boxes describe contextual (yellow shading) and latent (gray) factors that shape the result variables (white), alongside the study participants.}
    \label{fig:causal-graph}
\end{figure*}

\section{Experimental Details}
\label{sec:app:experimental_details}

\begin{figure*}[!t]
  \centering
  \begin{promptblock}
Help me generate a dataset of conversation starters. Each conversation takes place within a specific context, and the person is seeking a particular piece of information. The generated starters will be read aloud by human actors to simulate a real-life interaction. Follow these rules for the generation:

- Use a natural tone. We are emulating real-life mundane scenarios. 
- Assume the two people do not know each other.
- Make the starter's information dense. We aim to include named entities, numerical quantities, context-specific technical terms, and other relevant information. 
- The conversation starter MUST be 50-60 words long.
- Do not generate anything else beside the conversation starter.

Generate one conversation starter in English for this scenario following the scenario-specific instructions provided.

Scenario: 
One person enters a pharmacy and asks the counter staff for advice and medication. 

Instructions:
The person has to greet the pharmacist, explain their symptoms and recent medication history, and ask for advice and a new prescription. 
You can imagine the person can suffering from one of these generic illnesses: cold and flu symptoms, headaches and migraines, digestive issues, allergies, skin conditions, muscle and join pain, minor infections, sleep problems, eye and ear issues, menstrual and reproductive health.
Be sure to provide detailed descriptions of the symptoms, as well as the exact names and quantities of the medication.
  \end{promptblock}
  \caption{\textbf{Example prompt for synthetic data generation.} Over-the-counter pharmacy scenario.}
    \label{tab:prompt_generation}
\end{figure*}

\begin{figure*}[!t]
  \centering
  \begin{promptblock}
You are a professional English to Portuguese translator, tasked with providing translations suitable for use in Portugal. Your goal is to accurately convey the meaning and nuances of the original English text while adhering to Portuguese grammar, vocabulary, and cultural sensitivities.

Please translate the following English text into Portuguese, preserving the original formatting (e.g. double newlines, indentation). Produce only the Portuguese translation, without any additional explanations or commentary.
  \end{promptblock}
  \caption{\textbf{Prompt for Question Translation.} We use it to prepare Portuguese questions in \receiverchip{2}.}
    \label{tab:question_translation}
\end{figure*}

\begin{figure*}[!t]
  \centering
  \begin{promptblock}
You are an expert in quality assurance for speech translation, specializing in English-to-Portuguese for healthcare and everyday conversational scenarios. Your task is to generate question-answer pairs *in English* based on an original English passage. These QA pairs are designed to help a human evaluator quickly assess the accuracy of a Portuguese translation derived from the English audio/text.

The **Answers** to your questions should isolate **critical pieces of information** from the original passage. These are elements where a mistranslation into Portuguese would likely lead to significant misunderstanding, incorrect actions, or potential harm, especially in a healthcare context.

**Focus on extracting:**

1.  **Key Entities:**
    *   **Healthcare:** Specific symptoms (e.g., "sharp pain," "dizziness"), body parts, medications (names, dosages like "500 milligrams"), allergies, medical conditions, procedures.
    *   **Everyday:** Names of people/places, specific times, dates, quantities, objects.
2.  **Crucial Quantifiers and Qualifiers:** Numbers, amounts, frequencies (e.g., "twice a day," "three weeks"), severity levels (e.g., "mild," "severe"), durations.
3.  **Negations and Affirmations:** Clear "yes/no" type information, presence or absence of symptoms/conditions (e.g., "no fever," "is allergic").
4.  **Core Instructions/Actions:** Essential verbs and objects related to what someone needs to do (e.g., "take with food," "call the doctor," "meet at the corner").
5.  **Conditional Information:** Key details under specific circumstances (e.g., "if the pain worsens").

**The Questions should directly probe for these critical pieces of information.** The questions must be answerable *solely* from the provided English passage.

**Constraints:**

*   Both Questions and Answers MUST be in English.
*   Answers should be concise and represent the exact critical information.
*   Questions should be clear and unambiguous.
*   Generate a diverse set of questions covering different types of critical information present in the passage.
*   The goal is to create checks that quickly reveal if the core, essential information has been preserved in the Portuguese translation.

**Format:**

Q: <question1>
A: <answer1>

Q: <question2>
A: <answer2>

...

**Original Passage:**
{passage}

**Question-Answer Pairs:**
  \end{promptblock}
  \caption{\textbf{Prompt for Question Generation.} We use the generated questions in \receiverchip{2}.}
    \label{fig:prompt_questions}
\end{figure*}

\begin{figure}[!t]
  \centering
  \begin{promptblock}
Please convert this transcribed text into a fluent text. Preserve as much medical information as possible. The resulting text should contain at least 50-60 words.

\{seed\}
  \end{promptblock}

  \begin{promptblock}
Please convert this dialogue into a single turn request by the customer using all the details provided. The resulting text should be fluent text with at least 50-60 words.

\{seed\}
  \end{promptblock}
  
  \caption{\textbf{Paraphrasing Prompts.} Used to prepare the conversation starters from MT-MED (top) and BConTrasT (bottom). ``\texttt{seed}'' is the original data point from the respective dataset.}
    \label{tab:cleaning-prompts}
\end{figure}

\subsection{Starters Preparation}
\label{ssec:app:starters-preparation}

Our starters sourced from existing sources required some rephrasing to be transformed into a plausible conversation started. In practice, we used Gemini 2.5 Pro to conduct the rephrasing. In all cases, we sampled with temperature of 0.9, top\_p of 0.95, and top\_k of 20, and used the following system prompt: ``You are a language model specialized in generating high-quality synthetic data.''. Figure~\ref{tab:cleaning-prompts} reports the prompts used for both datasets.

\subsection{Validation of Conversation Starters and Questions}
\label{app:validation}

To assess the quality of our starters and associated questions, we conducted a manual validation on a sample of 20 starters associated with 10 sets of unique English questions (i.e. 7\% of the dataset, 20 starters and 200 questions). The analysis was conducted by one author with a background in translation studies. 
We verified the starter content and the grounding of such questions in the starter content. No issues were found with the starters. Out of 200 questions, only one raised an ambiguity, making it arguably unanswerable.
Such issues are thus rare and do not undermine the overall quality of the dataset. Moreover, they are accounted for in the validation process: receivers can explicitly flag a question as unanswerable, and validators can accept this response as correct.

\subsection{Scalar Quality Metric}
\label{sec:app:SQM}

For the quality assessment, we use the SQM scale that features seven labeled tick marks indicating different quality labels combining \textit{accuracy} and \textit{grammatical correctness} described as follows:

\begin{itemize}
    \item 6: Perfect Meaning and Grammar: The meaning of the translation is completely consistent with the source and the surrounding context (if applicable). The grammar is also correct. 
    \item 4: Most Meaning Preserved and Few Grammar Mistakes: The translation retains most of the meaning of the source. It may have some grammar mistakes or minor contextual inconsistencies. 
    \item 2: Some Meaning Preserved: The translation preserves some of the meaning of the source but misses significant parts. The narrative is hard to follow due to fundamental errors. Grammar may be poor. 
    \item 0: Nonsense/No meaning preserved: Nearly all information is lost between the translation and source. Grammar is irrelevant.
\end{itemize}

\subsection{Translation Details}
\label{app:ssec:translation-details}

We use Hugging Face's \texttt{transformers} classes and models to run all automatic translations \cite{wolf-etal-2020-transformers}. For Phi 4 (HF ID: microsoft/Phi-4-multimodal-instruct) and Voxtral Small (HF ID: mistralai/Voxtral-Small-24B-2507), we use the prompt ``Translate the audio to Portuguese.'', whereas for DeSTA2 we use the slightly more complex ``Translate this audio to Portuguese. Produce only the Portuguese translation, without any additional explanations or commentary.'' as the model showed the tendency to generate additional content besides the translation. We load Whisper v3 and Voxtral in bfloat16.
In all runs, we used a batch size of 1 and standard decoding parameters as implemented in \texttt{transformers}.
To translate the transcript with Tower+, we use the following prompt: ``Translate the following English source text to Portuguese (Portugal):\textbackslash nEnglish: \{text\}\textbackslash nPortuguese (Portugal): '' and greedy decoding.

\section{Additional Results}

\input{tab_starters-examples}

\subsection{Factor Analysis on Survey Scores}
\label{app:ssec:factor-analysis}

Since each sender reports three scores---satisfaction, trust, and reliance---we conduct a factor analysis to verify whether they represent distinct constructs or can be treated as a single composite measure. Given the nested structure of the data---10 interaction-level observations per participant---we decompose variance into a within-user component (person-mean-centered scores, $n = 1{,}200$) and a between-user component (per-user means, $n = 120$), and run exploratory factor analysis (EFA) independently on each. Prior to fitting, Bartlett's test of sphericity confirmed that the correlation matrices were significantly different from identity at both levels (within: $\chi^2 = 5927.9$, $p < .0001$; between: $\chi^2(3) = 733.6$, $p < .001$), and Kaiser-Meyer-Olkin (KMO) values indicated adequate sampling adequacy (within: $\text{KMO} = 0.781$; between: $\text{KMO} = 0.774$). Parallel analysis (500 iterations, 95th percentile threshold) retained a single factor at both levels. The one-factor solution explained $89.1\%$ of within-user variance and $92.7\%$ of between-user variance.
These results indicate that satisfaction, trust, and reliance do not function as empirically distinct constructs in our data: participants rated them nearly interchangeably across interactions and across individuals. We therefore average the three scores into a single composite measure, which serves as the dependent variable in all subsequent analyses.

\begin{figure}
    \centering
    \includegraphics[width=1\linewidth]{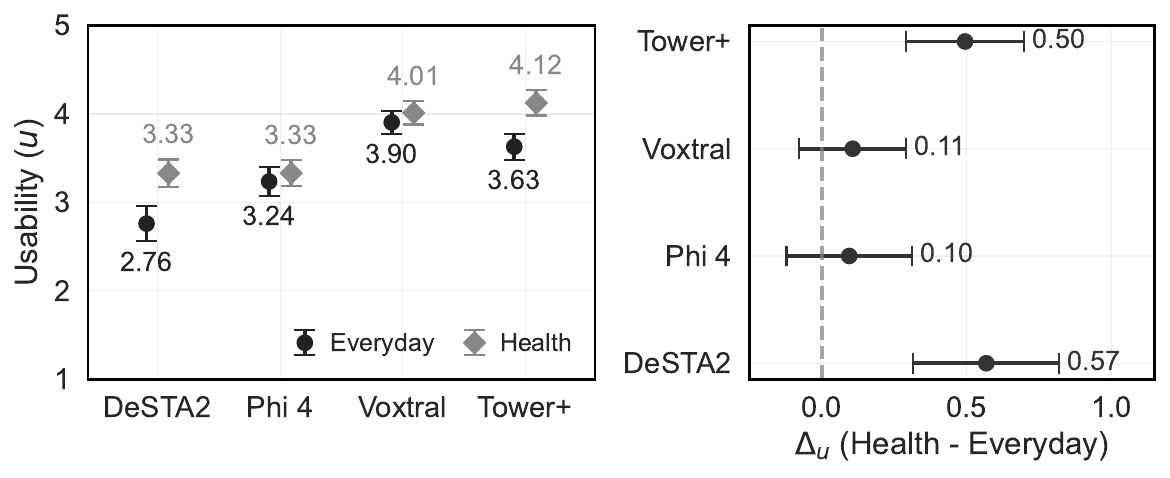}
    \caption{\textbf{Mean usability scores by translation model and conversation topic.} Left: estimated means $\pm$95\% CI for Everyday and Health conversations. Right: per-model difference $\Delta_u = u_\text{Health} - u_\text{Everyday}$ with 95\% CI; positive values indicate higher usability for health-related conversations.}
    \label{fig:usability-scores-translation-system}
\end{figure}

\begin{figure}[!t]
    \centering
    \includegraphics[width=1\linewidth]{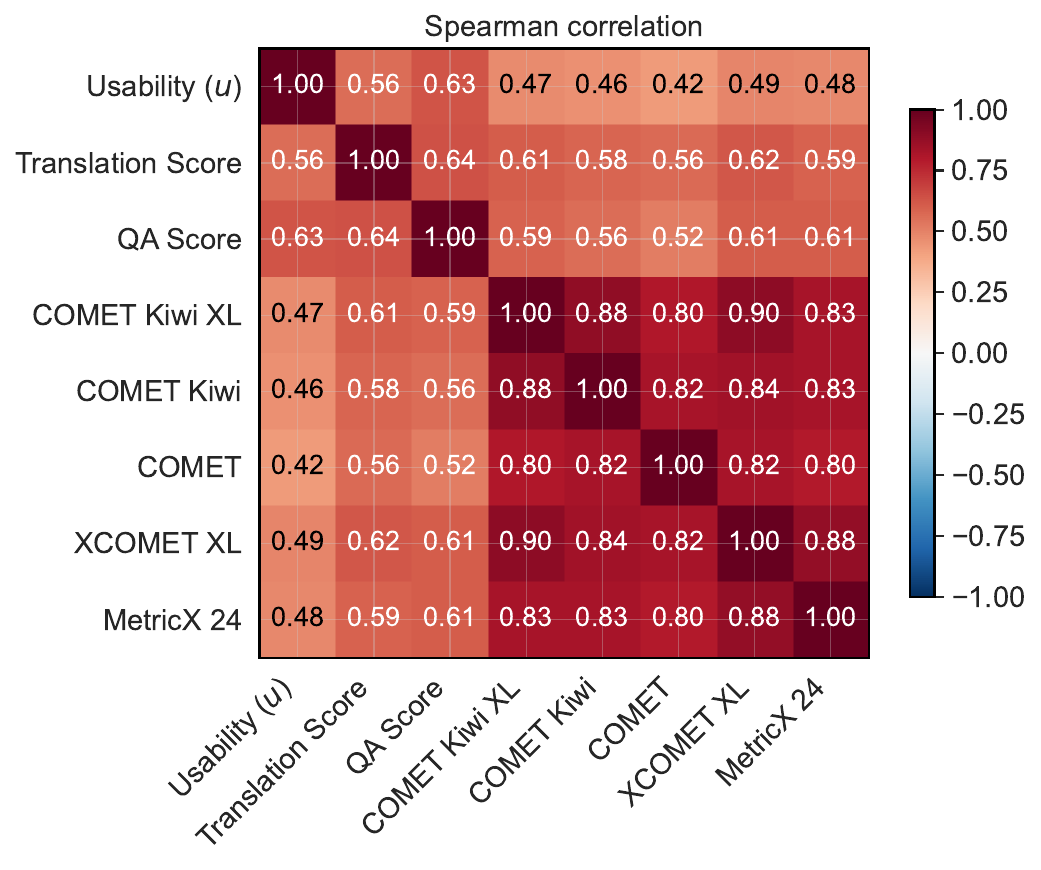}
    \caption{Spearman's $\rho$ correlation between Usability ($u$) and translation quality metrics.}
    \label{fig:metrics-correlation}
\end{figure}

\subsection{Linear Mixed Effects Model}
\label{app:ssec:lmm}

We use \texttt{pymer4} \cite{jolly2018pymer4} to fit Linear Mixed Effects Models in our analysis, using REML. Table~\ref{tab:lmer-coeffs} reports the coefficients, confidence intervals, and significance level ($p$) of the model used to predict the usability variable $u$.

\input{tab_lmer-coeffs}

\section{AI Use Statement}

We used AI coding tools to streamline the generation of visual artifacts of the paper, and writing assistants to polish parts of this manuscript.

\end{document}

%% file: commands.tex
\usepackage{tikz}
\usepackage{fvextra}

\definecolor{boxgray}{gray}{0.95}
\definecolor{sectionblue}{RGB}{0,102,204}

\definecolor{ouviagreen}{HTML}{778545}
\definecolor{ouviablue}{HTML}{5B8BC3}
\definecolor{ouviagold}{HTML}{DDA15E}

\DefineVerbatimEnvironment{promptblock}{Verbatim}{fontsize=\small,breaklines=true,commandchars=\\\{\},frame=single}

\newcommand{\softchipgreen}[2][ouviagreen]{%
  \tikz[baseline=(text.base)]{%
    \node[
      rounded corners=3pt,
      fill=#1,
      text=white,
      inner sep=3pt,
      font=\sffamily\small
    ] (text) {#2};%
  }%
  \xspace%
}
\newcommand{\softchipblue}[2][ouviablue]{%
  \tikz[baseline=(text.base)]{%
    \node[
      rounded corners=3pt,
      fill=#1,
      text=white,
      inner sep=3pt,
      font=\sffamily\small
    ] (text) {#2};%
  }%
  \xspace%
}
\newcommand{\softchipgold}[2][ouviagold]{%
  \tikz[baseline=(text.base)]{%
    \node[
      rounded corners=3pt,
      fill=#1,
      text=white,
      inner sep=3pt,
      font=\sffamily\small
    ] (text) {#2};%
  }%
  \xspace%
}

\newcommand{\senderchip}{\softchipgreen}
\newcommand{\receiverchip}{\softchipblue}
\newcommand{\validatorchip}{\softchipgold}

%% file: tab_scores.tex
\begin{table}[!t]
\centering
\small
\setlength{\tabcolsep}{4pt}
\begin{tabular}{@{}p{1cm}rrrrrr@{}}
\toprule
\textbf{} &
  \multicolumn{1}{c}{\textbf{S}} &
  \multicolumn{1}{c}{\textbf{T}} &
  \multicolumn{1}{c}{\textbf{R}} &
  \multicolumn{1}{c}{\textbf{$\mu(u)$}} &
  \multicolumn{1}{c}{\textbf{$\sigma(u)$}} &
  \multicolumn{1}{c}{\textbf{$\Delta_u$}} \\ \midrule
US\textsubscript{W} &
  \multicolumn{1}{r}{\textbf{3.91}} &
  \multicolumn{1}{r}{\textbf{3.84}} &
  \multicolumn{1}{r}{\textbf{3.86}} &
  \multicolumn{1}{r}{\textbf{3.87}} &
  \multicolumn{1}{r}{1.15} &
  \multicolumn{1}{r}{-0.19} \\
~~{\small $\text{b}\rfloor$} &
  {\small 4.11} &
  {\small 4.05} &
  {\small 4.01} &
  {\small 4.06} &
  {\small 0.86} &
   \\ \addlinespace[2pt]

US\textsubscript{B} &
  \multicolumn{1}{r}{3.50} &
  \multicolumn{1}{r}{3.48} &
  \multicolumn{1}{r}{3.46} &
  \multicolumn{1}{r}{3.48} &
  \multicolumn{1}{r}{1.18} &
  \multicolumn{1}{r}{-0.20} \\ 
~~{\small $\text{b}\rfloor$} &
  {\small 3.7} &
  {\small 3.66} &
  {\small 3.69} &
  {\small 3.68} &
  {\small 0.91} &
   \\ \addlinespace[2pt]

Hindi &
  \multicolumn{1}{r}{3.40} &
  \multicolumn{1}{r}{3.36} &
  \multicolumn{1}{r}{3.31} &
  \multicolumn{1}{r}{3.35} &
  \multicolumn{1}{r}{1.20} &
  \multicolumn{1}{r}{-0.40} \\
~~{\small $\text{b}\rfloor$} &
  {\small 3.78} &
  {\small 3.74} &
  {\small 3.75} &
  {\small 3.76} &
  {\small 0.88} &
   \\ \bottomrule
\end{tabular}
\caption{\textbf{5-point Likert usability scores by language group.} \textbf{S}atisfaction, \textbf{T}rust, and \textbf{R}eliance are averaged into a composite score $u$; we report the group mean $\mu(u)$, standard deviation $\sigma(u)$, and mean shift relative to $u$ in the baseline condition ($\Delta_u$). Baseline {\small ($\text{b}\rfloor$)} scores beneath each group.}
\label{tab:survey-scores}
\end{table}

%% file: fig_usability_drilldown.tex
\begin{figure*}[!t]
  \centering
    \begin{subfigure}[b]{0.34\linewidth}
        \centering
        \includegraphics[width=\textwidth]{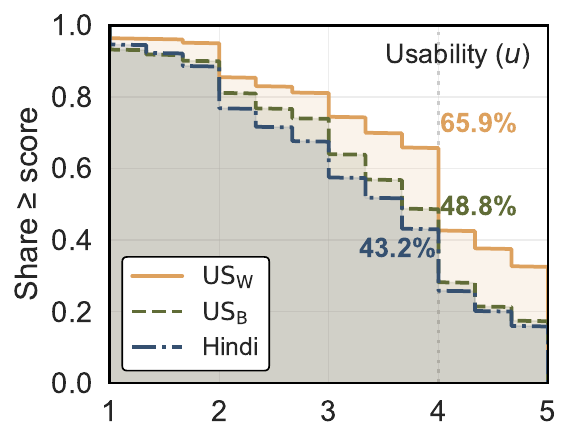}
        \caption{}
        \label{fig:usability-score-distribution-group}
    \end{subfigure}
    \hfill
  \begin{subfigure}[b]{0.32\linewidth}
        \centering
        \includegraphics[width=\textwidth]{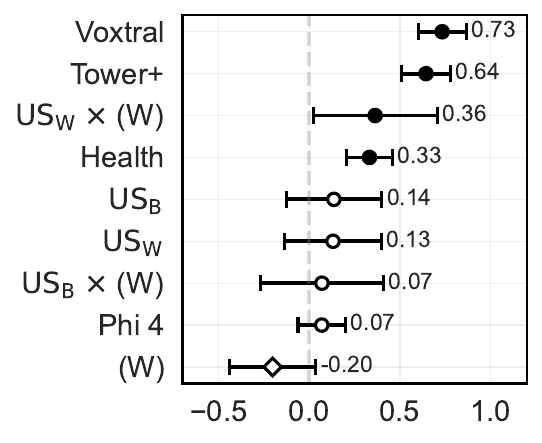}
        \caption{}
        \label{fig:usability-coeff}
    \end{subfigure}
    \begin{subfigure}[b]{0.29\linewidth}
        \centering
        \includegraphics[width=\textwidth]{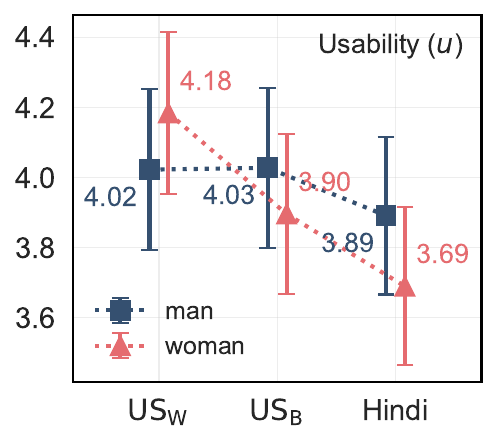}
        \caption{}
        \label{fig:usability-interaction-group-gender}
    \end{subfigure}

  \caption{
  (\textbf{a)} Survival curves of usability scores by language group. Each curve shows the fraction of the group whose usability rating reaches or exceeds a given score on the x-axis. Annotations at $u=4$.
  \textbf{(b)} Fixed-effect estimates (95\% CI) from a LMM predicting usability ($u$) (\S\ref{ssec:findings-usability}). Filled circles $p < 0.05$, diamonds $0.05 \leq p < 0.1$, hollow non-significant. Estimates annotated right of each interval.
  \textbf{(c)} Estimated marginal means ($\pm$95\% CI) of usability by language group and gender, marginalized over topic and model.}
  \label{fig:usability-drilldown}
\end{figure*}

%% file: fig_validator-translation-score.tex
\begin{figure*}[!t]
  \centering
    \begin{subfigure}[b]{0.3\textwidth}
        \centering
        \includegraphics[width=\textwidth]{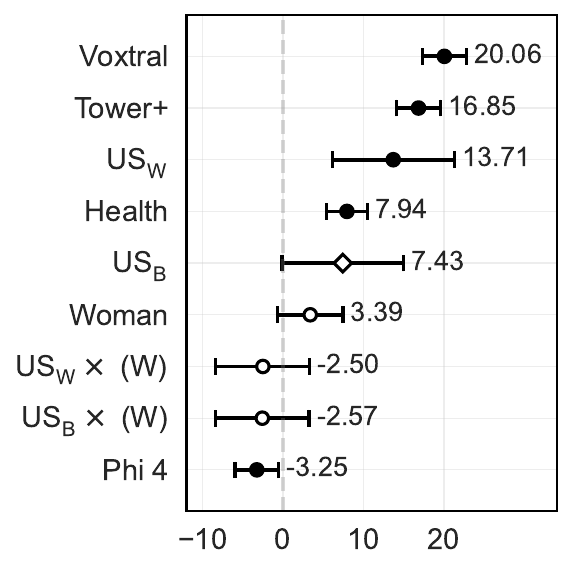}
        \caption{}
        \label{fig:vs-coeff}
    \end{subfigure}
    \begin{subfigure}[b]{0.3\textwidth}
        \centering
        \includegraphics[width=\textwidth]{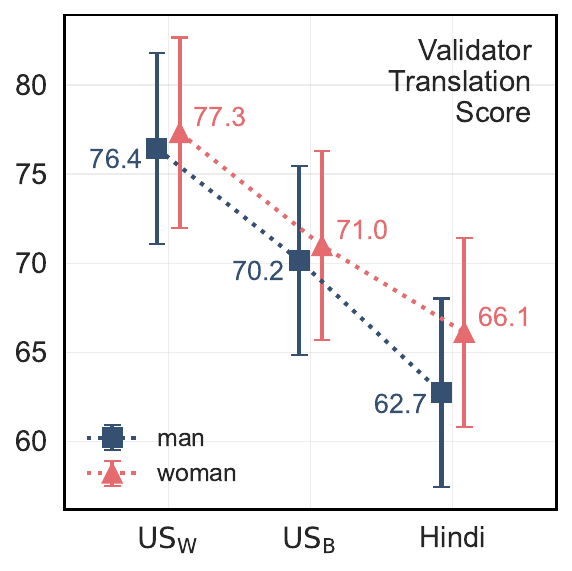}
        \caption{}
        \label{fig:vs-group-gender}
    \end{subfigure}
    \hfill
  \begin{subfigure}[b]{0.31\textwidth}
        \centering
        \includegraphics[width=\textwidth]{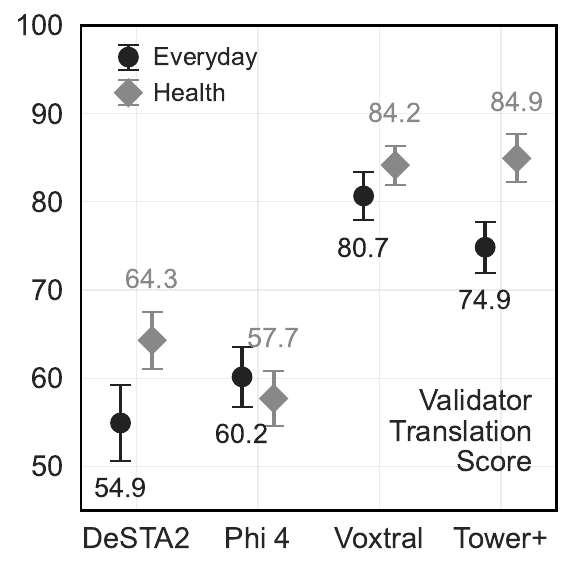}
        \caption{}
        \label{fig:vs-model}
    \end{subfigure}

  \caption{
    \textbf{(a)} Fixed-effect estimates (95\% CI) from a linear mixed model predicting validator
    translation score. Reference: Hindi, Man, DeSTA2, Everyday, MED-MT. Filled
    circles $p<0.05$, diamonds $0.05 \leq p<0.1$, hollow non-significant.
    \textbf{(b)} Model-adjusted marginal means ($\pm$95\% CI) by language group and gender,
    marginalized over topic, model, and source.
    \textbf{(c)} Descriptive means ($\pm$95\% CI) by translation model and conversation topic.}
  \label{fig:validator-score}
\end{figure*}

%% file: tab_starters-examples.tex
\begin{table*}[!t]
    \small
    \setlength{\tabcolsep}{4pt}
    \centering
    \begin{tabular}{lp{1.8cm}p{0.67\textwidth}c}
        \toprule
        \textbf{Source} & \textbf{Topic} & \textbf{Example} & \textbf{Count} \\ \midrule
        MED-MT & Patient-Physician & Hello, I came in today because I have been feeling unwell for about a week now. My primary symptom is a very persistent and sore throat that hasn't improved. In addition to that, I have also started experiencing chills over the last few nights, so I thought it would be best to get it checked out professionally. & 50 \\ \addlinespace
        Synthetic & Over-the-counter Pharmacy & Hi there, I'm hoping you can help. For the past five days, I've had terrible seasonal allergies—constant sneezing, a runny nose, and really itchy, watery eyes. I've been taking one 10mg Cetirizine tablet every morning, but it's barely making a dent. Is there a more effective over-the-counter option you'd recommend for these symptoms? & 100 \\ \addlinespace \midrule \midrule \addlinespace
        BConTrasT & Customer Support & Hello, I'd like to place an order for pickup from Bella Luna Pizzeria. I need four large pizzas, all with extra cheese, please. The first one should have pepperoni and pineapple, the second one should be the five-cheese blend, the third a taco pizza, and the last one should just have tomatoes, olives, and green peas. & 50 \\ \addlinespace
        Synthetic & Everyday & Excuse me, good morning. I'm trying to get to the Belém Tower from here at Rossio station. I believe I need to take the 15E tram, is that correct? I'll need to buy a Viva Viagem card for two people for a single trip; could you tell me the total cost and the approximate travel time around 11 AM? & 100 \\ \bottomrule
    \end{tabular}
    \caption{\textbf{Conversation starters.} These prompts simulate a real-life beginning of conversation in high- (top) and low- (bottom) stakes scenarios. Synthetic rows are new. Absolute counts per data source.}
    \label{tab:starters-examples}
\end{table*}

%% file: tab_lmer-coeffs.tex
\begin{table*}[!t]
\small
\centering
\begin{tabular}{@{}lrrrrrr@{}}
\toprule
\textbf{term} &
  \multicolumn{1}{c}{\textbf{Estimate}} &
  \multicolumn{1}{c}{\textbf{Std. Error}} &
  \multicolumn{1}{c}{\textbf{t}} &
  \multicolumn{1}{c}{\textbf{p}} &
  \multicolumn{1}{c}{\textbf{2.5\%}} &
  \multicolumn{1}{c}{\textbf{97.5\%}} \\ \midrule
(Intercept) & 1.564 & 0.164 & 9.535 & 0.000 & 1.242 & 1.886 \\
Group: US\textsubscript{B} & 0.136 & 0.133 & 1.022 & 0.308 & -0.127 & 0.399 \\
Group: US\textsubscript{W} & 0.131 & 0.135 & 0.975 & 0.331 & -0.135 & 0.397 \\
Gender: Woman & -0.202 & 0.120 & -1.677 & 0.096 & -0.440 & 0.036 \\
Model: Phi 4 & 0.069 & 0.067 & 1.035 & 0.301 & -0.062 & 0.200 \\
Model: Tower+ & 0.644 & 0.069 & 9.378 & 0.000 & 0.509 & 0.779 \\
Model: Voxtral & 0.733 & 0.068 & 10.860 & 0.000 & 0.601 & 0.865 \\
$u$ (Baseline) & 0.466 & 0.030 & 15.504 & 0.000 & 0.407 & 0.525 \\
Topic: Health & 0.332 & 0.064 & 5.177 & 0.000 & 0.206 & 0.459 \\
Source: Gemini & -0.541 & 0.084 & -6.483 & 0.000 & -0.706 & -0.377 \\
Source: MED-MT & -0.343 & 0.109 & -3.147 & 0.002 & -0.557 & -0.128 \\
Source: Qwen 3 & -0.452 & 0.083 & -5.413 & 0.000 & -0.616 & -0.287 \\
Group: US\textsubscript{B} × Gender: Woman & 0.070 & 0.172 & 0.408 & 0.684 & -0.269 & 0.410 \\
Group: US\textsubscript{W} × Gender: Woman & 0.363 & 0.173 & 2.098 & 0.038 & 0.021 & 0.706 \\ \bottomrule
\end{tabular}
\caption{\textbf{Fixed-effect estimates.} Linear mixed-effects model predicting usability ($u$) across $N=1{,}738$ observations. The model includes group, gender, and their interaction as focal predictors, with conversation topic, translation model, baseline usability, validator translation score, and correction ratio as covariates. Random intercepts are included for participants and conversations. The reference category is Hindi, Man,  DeSTA2, BConTrasT, Everyday. CIs are 95\% Wald intervals.}
\label{tab:lmer-coeffs}
\end{table*}

%% file: minimal.bib
@proceedings{iwslt-ws-2025-1,
  address =       {Vienna, Austria (in-person and online)},
  editor =        {Salesky, Elizabeth and Federico, Marcello and
                   Anastasopoulos, Antonis},
  month =         jul,
  publisher =     {Association for Computational Linguistics},
  title =         {Proceedings of the 22nd International Conference on
                   Spoken Language Translation (IWSLT 2025)},
  year =          {2025},
  doi =           {10.18653/v1/2025.iwslt-1.0},
  isbn =          {979-8-89176-272-5},
  url =           {https://aclanthology.org/2025.iwslt-1.0/},
}

@inproceedings{rei-etal-2022-comet,
  address =       {Abu Dhabi, United Arab Emirates (Hybrid)},
  author =        {Rei, Ricardo and C. de Souza, Jos{\'e} G. and
                   Alves, Duarte and Zerva, Chrysoula and Farinha, Ana C and
                   Glushkova, Taisiya and Lavie, Alon and Coheur, Luisa and
                   Martins, Andr{\'e} F. T.},
  booktitle =     {Proceedings of the Seventh Conference on Machine
                   Translation (WMT)},
  editor =        {Koehn, Philipp and Barrault, Lo{\"i}c and
                   Bojar, Ond{\v{r}}ej and Bougares, Fethi and
                   Chatterjee, Rajen and Costa-juss{\`a}, Marta R. and
                   Federmann, Christian and Fishel, Mark and
                   Fraser, Alexander and Freitag, Markus and
                   Graham, Yvette and Grundkiewicz, Roman and
                   Guzman, Paco and Haddow, Barry and Huck, Matthias and
                   Jimeno Yepes, Antonio and Kocmi, Tom and
                   Martins, Andr{\'e} and Morishita, Makoto and
                   Monz, Christof and Nagata, Masaaki and
                   Nakazawa, Toshiaki and Negri, Matteo and
                   N{\'e}v{\'e}ol, Aur{\'e}lie and Neves, Mariana and
                   Popel, Martin and Turchi, Marco and Zampieri, Marcos},
  month =         dec,
  pages =         {578--585},
  publisher =     {Association for Computational Linguistics},
  title =         {{COMET}-22: Unbabel-{IST} 2022 Submission for the
                   Metrics Shared Task},
  year =          {2022},
  doi =           {10.18653/v1/2022.wmt-1.52},
  url =           {https://aclanthology.org/2022.wmt-1.52/},
}

@inproceedings{han-etal-2024-speechqe,
  address =       {Miami, Florida, USA},
  author =        {Han, HyoJung and Duh, Kevin and Carpuat, Marine},
  booktitle =     {Proceedings of the 2024 Conference on Empirical
                   Methods in Natural Language Processing},
  editor =        {Al-Onaizan, Yaser and Bansal, Mohit and
                   Chen, Yun-Nung},
  month =         nov,
  pages =         {21852--21867},
  publisher =     {Association for Computational Linguistics},
  title =         {{S}peech{QE}: Estimating the Quality of Direct Speech
                   Translation},
  year =          {2024},
  doi =           {10.18653/v1/2024.emnlp-main.1218},
  url =           {https://aclanthology.org/2024.emnlp-main.1218/},
}

@inproceedings{juraska-etal-2025-metricx,
  address =       {Suzhou, China},
  author =        {Juraska, Juraj and Domhan, Tobias and
                   Finkelstein, Mara and Nakagawa, Tetsuji and
                   Kovacs, Geza and Deutsch, Daniel and Wang, Pidong and
                   Freitag, Markus},
  booktitle =     {Proceedings of the Tenth Conference on Machine
                   Translation},
  editor =        {Haddow, Barry and Kocmi, Tom and Koehn, Philipp and
                   Monz, Christof},
  month =         nov,
  pages =         {957--968},
  publisher =     {Association for Computational Linguistics},
  title =         {{M}etric{X}-25 and {G}em{S}pan{E}val: {G}oogle
                   {T}ranslate Submissions to the {WMT}25 Evaluation
                   Shared Task},
  year =          {2025},
  doi =           {10.18653/v1/2025.wmt-1.70},
  isbn =          {979-8-89176-341-8},
  url =           {https://aclanthology.org/2025.wmt-1.70/},
}

@article{colina2009further,
  author =        {Colina, Sonia},
  journal =       {Target. International Journal of Translation Studies},
  number =        {2},
  pages =         {235--264},
  publisher =     {John Benjamins},
  title =         {Further evidence for a functionalist approach to
                   translation quality evaluation},
  volume =        {21},
  year =          {2009},
}

@inproceedings{liu-etal-2024-evaluation,
  address =       {Chicago, USA},
  author =        {Liu, Ting and Lo, Chi-kiu and Marshman, Elizabeth and
                   Knowles, Rebecca},
  booktitle =     {Proceedings of the 16th Conference of the Association
                   for Machine Translation in the Americas (Volume 1:
                   Research Track)},
  editor =        {Knowles, Rebecca and Eriguchi, Akiko and
                   Goel, Shivali},
  month =         sep,
  pages =         {190--208},
  publisher =     {Association for Machine Translation in the Americas},
  title =         {Evaluation Briefs: Drawing on Translation Studies for
                   Human Evaluation of {MT}},
  year =          {2024},
  url =           {https://aclanthology.org/2024.amta-research.17/},
}

@article{khoong2019assessing,
  author =        {Khoong, Elaine C and Steinbrook, Eric and
                   Brown, Cortlyn and Fernandez, Alicia},
  journal =       {JAMA internal medicine},
  number =        {4},
  pages =         {580--582},
  title =         {Assessing the use of Google Translate for Spanish and
                   Chinese translations of emergency department
                   discharge instructions},
  volume =        {179},
  year =          {2019},
}

@inproceedings{savoldi-etal-2024-harm,
  address =       {Miami, Florida, USA},
  author =        {Savoldi, Beatrice and Papi, Sara and Negri, Matteo and
                   Guerberof-Arenas, Ana and Bentivogli, Luisa},
  booktitle =     {Proceedings of the 2024 Conference on Empirical
                   Methods in Natural Language Processing},
  editor =        {Al-Onaizan, Yaser and Bansal, Mohit and
                   Chen, Yun-Nung},
  month =         nov,
  pages =         {18048--18076},
  publisher =     {Association for Computational Linguistics},
  title =         {What the Harm? Quantifying the Tangible Impact of
                   Gender Bias in Machine Translation with a
                   Human-centered Study},
  year =          {2024},
  doi =           {10.18653/v1/2024.emnlp-main.1002},
  url =           {https://aclanthology.org/2024.emnlp-main.1002/},
}

@inproceedings{zaranis-etal-2025-watching,
  address =       {Vienna, Austria},
  author =        {Zaranis, Emmanouil and Attanasio, Giuseppe and
                   Agrawal, Sweta and Martins, Andre},
  booktitle =     {Proceedings of the 63rd Annual Meeting of the
                   Association for Computational Linguistics (Volume 1:
                   Long Papers)},
  editor =        {Che, Wanxiang and Nabende, Joyce and
                   Shutova, Ekaterina and Pilehvar, Mohammad Taher},
  month =         jul,
  pages =         {25261--25284},
  publisher =     {Association for Computational Linguistics},
  title =         {Watching the Watchers: Exposing Gender Disparities in
                   Machine Translation Quality Estimation},
  year =          {2025},
  doi =           {10.18653/v1/2025.acl-long.1228},
  isbn =          {979-8-89176-251-0},
  url =           {https://aclanthology.org/2025.acl-long.1228/},
}

@article{nurminen2021investigating,
  author =        {Nurminen, Mary},
  publisher =     {Tampere University},
  title =         {Investigating the influence of context in the use and
                   reception of raw machine translation},
  year =          {2021},
}

@inproceedings{savoldi-etal-2025-translation,
  address =       {Suzhou, China},
  author =        {Savoldi, Beatrice and Ramponi, Alan and Negri, Matteo and
                   Bentivogli, Luisa},
  booktitle =     {Proceedings of the 2025 Conference on Empirical
                   Methods in Natural Language Processing},
  editor =        {Christodoulopoulos, Christos and Chakraborty, Tanmoy and
                   Rose, Carolyn and Peng, Violet},
  month =         nov,
  pages =         {13876--13889},
  publisher =     {Association for Computational Linguistics},
  title =         {Translation in the Hands of Many: Centering Lay Users
                   in Machine Translation Interactions},
  year =          {2025},
  doi =           {10.18653/v1/2025.emnlp-main.700},
  isbn =          {979-8-89176-332-6},
  url =           {https://aclanthology.org/2025.emnlp-main.700/},
}

@inproceedings{carpuat-etal-2025-interdisciplinary,
  address =       {Suzhou, China},
  author =        {Carpuat, Marine and Asscher, Omri and Bali, Kalika and
                   Bentivogli, Luisa and Blain, Fr{\'e}d{\'e}ric and
                   Bowker, Lynne and Choudhury, Monojit and
                   Daum{\'e} III, Hal and Duh, Kevin and Gao, Ge and
                   Grissom II, Alvin and Karpinska, Marzena and
                   Khoong, Elaine C. and Lewis, William D. and
                   Martins, Andr{\'e} F. T. and Nurminen, Mary and
                   Oard, Douglas W. and Popovic, Maja and Simard, Michel and
                   Yvon, Fran{\c{c}}ois},
  booktitle =     {Proceedings of the 2025 Conference on Empirical
                   Methods in Natural Language Processing},
  editor =        {Christodoulopoulos, Christos and Chakraborty, Tanmoy and
                   Rose, Carolyn and Peng, Violet},
  month =         nov,
  pages =         {22859--22879},
  publisher =     {Association for Computational Linguistics},
  title =         {An Interdisciplinary Approach to Human-Centered
                   Machine Translation},
  year =          {2025},
  doi =           {10.18653/v1/2025.emnlp-main.1164},
  isbn =          {979-8-89176-332-6},
  url =           {https://aclanthology.org/2025.emnlp-main.1164/},
}

@techreport{aima2025,
  address =       {Lisboa},
  author =        {{AIMA}},
  institution =   {Agência para a Integração, Migrações e Asilo
                   (AIMA I.P.)},
  month =         {10},
  note =          {Edição Digital. Coordenação: Sílvia Lopes},
  title =         {Relatório de Migrações e Asilo 2024},
  year =          {2025},
  isbn =          {978-989-35625-8-1},
  url =           {https://aima.gov.pt/media/pages/documents/fec4d6a712-
                  1760603125/relatorio-migracoes-e-asilo-2024.pdf},
}

@article{doi:10.1073/pnas.1915768117,
  author =        {Allison Koenecke and Andrew Nam and Emily Lake and
                   Joe Nudell and Minnie Quartey and Zion Mengesha and
                   Connor Toups and John R. Rickford and Dan Jurafsky and
                   Sharad Goel},
  journal =       {Proceedings of the National Academy of Sciences},
  number =        {14},
  pages =         {7684-7689},
  title =         {Racial disparities in automated speech recognition},
  volume =        {117},
  year =          {2020},
  doi =           {10.1073/pnas.1915768117},
  url =           {https://www.pnas.org/doi/abs/10.1073/pnas.1915768117},
}

@article{hovy2002principles,
  author =        {Hovy, Eduard and King, Margaret and
                   Popescu-Belis, Andrei},
  journal =       {Machine Translation},
  number =        {1},
  pages =         {43--75},
  publisher =     {Springer},
  title =         {Principles of context-based machine translation
                   evaluation},
  volume =        {17},
  year =          {2002},
}

@article{hutchins2005current,
  author =        {Hutchins, John},
  journal =       {International journal of translation},
  number =        {1-2},
  pages =         {5--38},
  title =         {Current commercial machine translation systems and
                   computer-based translation tools: system types and
                   their uses},
  volume =        {17},
  year =          {2005},
}

@inproceedings{ai-etal-2007-comparing,
  address =       {Antwerp, Belgium},
  author =        {Ai, Hua and Raux, Antoine and Bohus, Dan and
                   Eskenazi, Maxine and Litman, Diane},
  booktitle =     {Proceedings of the 8th SIGdial Workshop on Discourse
                   and Dialogue},
  editor =        {Bunt, Harry and Keizer, Simon and Paek, Tim},
  month =         sep,
  pages =         {124--131},
  publisher =     {Association for Computational Linguistics},
  title =         {Comparing Spoken Dialog Corpora Collected with
                   Recruited Subjects versus Real Users},
  year =          {2007},
  doi =           {10.18653/v1/2007.sigdial-1.23},
  url =           {https://aclanthology.org/2007.sigdial-1.23/},
}

@inproceedings{graham-etal-2013-continuous,
  address =       {Sofia, Bulgaria},
  author =        {Graham, Yvette and Baldwin, Timothy and
                   Moffat, Alistair and Zobel, Justin},
  booktitle =     {Proceedings of the 7th Linguistic Annotation Workshop
                   and Interoperability with Discourse},
  editor =        {Pareja-Lora, Antonio and Liakata, Maria and
                   Dipper, Stefanie},
  month =         aug,
  pages =         {33--41},
  publisher =     {Association for Computational Linguistics},
  title =         {Continuous Measurement Scales in Human Evaluation of
                   Machine Translation},
  year =          {2013},
  url =           {https://aclanthology.org/W13-2305/},
}

@inproceedings{kocmi-etal-2022-findings,
  address =       {Abu Dhabi, United Arab Emirates (Hybrid)},
  author =        {Kocmi, Tom and Bawden, Rachel and Bojar, Ond{\v{r}}ej and
                   Dvorkovich, Anton and Federmann, Christian and
                   Fishel, Mark and Gowda, Thamme and Graham, Yvette and
                   Grundkiewicz, Roman and Haddow, Barry and
                   Knowles, Rebecca and Koehn, Philipp and
                   Monz, Christof and Morishita, Makoto and
                   Nagata, Masaaki and Nakazawa, Toshiaki and
                   Nov{\'a}k, Michal and Popel, Martin and
                   Popovi{\'c}, Maja},
  booktitle =     {Proceedings of the Seventh Conference on Machine
                   Translation (WMT)},
  editor =        {Koehn, Philipp and Barrault, Lo{\"i}c and
                   Bojar, Ond{\v{r}}ej and Bougares, Fethi and
                   Chatterjee, Rajen and Costa-juss{\`a}, Marta R. and
                   Federmann, Christian and Fishel, Mark and
                   Fraser, Alexander and Freitag, Markus and
                   Graham, Yvette and Grundkiewicz, Roman and
                   Guzman, Paco and Haddow, Barry and Huck, Matthias and
                   Jimeno Yepes, Antonio and Kocmi, Tom and
                   Martins, Andr{\'e} and Morishita, Makoto and
                   Monz, Christof and Nagata, Masaaki and
                   Nakazawa, Toshiaki and Negri, Matteo and
                   N{\'e}v{\'e}ol, Aur{\'e}lie and Neves, Mariana and
                   Popel, Martin and Turchi, Marco and Zampieri, Marcos},
  month =         dec,
  pages =         {1--45},
  publisher =     {Association for Computational Linguistics},
  title =         {Findings of the 2022 Conference on Machine
                   Translation ({WMT}22)},
  year =          {2022},
  doi =           {10.18653/v1/2022.wmt-1.1},
  url =           {https://aclanthology.org/2022.wmt-1.1/},
}

@inproceedings{basoah2025not,
  author =        {Basoah, Jeffrey and Chechelnitsky, Daniel and
                   Long, Tao and Reinecke, Katharina and
                   Zerva, Chrysoula and Zhou, Kaitlyn and
                   D{\'\i}az, Mark and Sap, Maarten},
  booktitle =     {Proceedings of the 2025 ACM Conference on Fairness,
                   Accountability, and Transparency},
  pages =         {710--745},
  title =         {Not like us, hunty: Measuring perceptions and
                   behavioral effects of minoritized anthropomorphic
                   cues in llms},
  year =          {2025},
}

@article{hoffman-trust,
  author =        {Hoffman, Robert R. and Mueller, Shane T. and
                   Klein, Gary and Litman, Jordan},
  journal =       {Frontiers in Computer Science},
  title =         {Measures for explainable AI: Explanation goodness,
                   user satisfaction, mental models, curiosity, trust,
                   and human-AI performance},
  volume =        {Volume 5 - 2023},
  year =          {2023},
  doi =           {10.3389/fcomp.2023.1096257},
  issn =          {2624-9898},
  url =           {https://www.frontiersin.org/journals/computer-science/
                  articles/10.3389/fcomp.2023.1096257},
}

@article{McGrath-trust-reliance,
  author =        {McGrath, Melanie J. and Lack, Oliver and Tisch, James and
                   Duenser, Andreas},
  journal =       {Frontiers in Artificial Intelligence},
  title =         {Measuring trust in artificial intelligence:
                   validation of an established scale and its short
                   form},
  volume =        {Volume 8 - 2025},
  year =          {2025},
  doi =           {10.3389/frai.2025.1582880},
  issn =          {2624-8212},
  url =           {https://www.frontiersin.org/journals/artificial-
                  intelligence/articles/10.3389/frai.2025.1582880},
}

@techreport{iso9241-11-2018,
  author =        {{ISO}},
  institution =   {International Organization for Standardization},
  type =          {Standard},
  title =         {{ISO} 9241-11:2018 --- Ergonomics of Human-System
                   Interaction---Part 11: Usability: Definitions and
                   Concepts},
  year =          {2018},
  url =           {https://www.iso.org/obp/ui/en/#iso:std:iso:9241:-11:ed-2:v1:
                  en},
}

@article{genovese2024artificial,
  author =        {Genovese, Ariana and Borna, Sahar and
                   Gomez-Cabello, Cesar A and Haider, Syed Ali and
                   Prabha, Srinivasagam and Forte, Antonio J and
                   Veenstra, Benjamin R},
  journal =       {Annals of Translational Medicine},
  number =        {6},
  pages =         {117},
  publisher =     {LWW},
  title =         {Artificial intelligence in clinical settings: a
                   systematic review of its role in language translation
                   and interpretation},
  volume =        {12},
  year =          {2024},
}

@article{taira2021pragmatic,
  author =        {Taira, Breena R and Kreger, Vanessa and
                   Orue, Aristides and Diamond, Lisa C},
  journal =       {Journal of General Internal Medicine},
  number =        {11},
  pages =         {3361--3365},
  publisher =     {Springer},
  title =         {A pragmatic assessment of Google Translate for
                   emergency department instructions},
  volume =        {36},
  year =          {2021},
}

@article{fareez2022dataset,
  author =        {Fareez, Faiha and Parikh, Tishya and
                   Wavell, Christopher and Shahab, Saba and
                   Chevalier, Meghan and Good, Scott and
                   De Blasi, Isabella and Rhouma, Rafik and
                   McMahon, Christopher and Lam, Jean-Paul and others},
  journal =       {Scientific Data},
  number =        {1},
  pages =         {313},
  publisher =     {Nature Publishing Group UK London},
  title =         {A dataset of simulated patient-physician medical
                   interviews with a focus on respiratory cases},
  volume =        {9},
  year =          {2022},
}

@inproceedings{farajian-etal-2020-findings,
  address =       {Online},
  author =        {Farajian, M. Amin and Lopes, Ant{\'o}nio V. and
                   Martins, Andr{\'e} F. T. and Maruf, Sameen and
                   Haffari, Gholamreza},
  booktitle =     {Proceedings of the Fifth Conference on Machine
                   Translation},
  editor =        {Barrault, Lo{\"i}c and Bojar, Ond{\v{r}}ej and
                   Bougares, Fethi and Chatterjee, Rajen and
                   Costa-juss{\`a}, Marta R. and Federmann, Christian and
                   Fishel, Mark and Fraser, Alexander and Graham, Yvette and
                   Guzman, Paco and Haddow, Barry and Huck, Matthias and
                   Yepes, Antonio Jimeno and Koehn, Philipp and
                   Martins, Andr{\'e} and Morishita, Makoto and
                   Monz, Christof and Nagata, Masaaki and
                   Nakazawa, Toshiaki and Negri, Matteo},
  month =         nov,
  pages =         {65--75},
  publisher =     {Association for Computational Linguistics},
  title =         {Findings of the {WMT} 2020 Shared Task on Chat
                   Translation},
  year =          {2020},
  doi =           {10.18653/v1/2020.wmt-1.3},
  url =           {https://aclanthology.org/2020.wmt-1.3/},
}

@inproceedings{ki-etal-2025-askqe,
  address =       {Vienna, Austria},
  author =        {Ki, Dayeon and Duh, Kevin and Carpuat, Marine},
  booktitle =     {Findings of the Association for Computational
                   Linguistics: ACL 2025},
  editor =        {Che, Wanxiang and Nabende, Joyce and
                   Shutova, Ekaterina and Pilehvar, Mohammad Taher},
  month =         jul,
  pages =         {17478--17515},
  publisher =     {Association for Computational Linguistics},
  title =         {{A}sk{QE}: Question Answering as Automatic Evaluation
                   for Machine Translation},
  year =          {2025},
  doi =           {10.18653/v1/2025.findings-acl.899},
  isbn =          {979-8-89176-256-5},
  url =           {https://aclanthology.org/2025.findings-acl.899/},
}

@article{fernandes2025llms,
  author =        {Fernandes, Patrick and Agrawal, Sweta and
                   Zaranis, Emmanouil and Martins, Andr{\'e} FT and
                   Neubig, Graham},
  journal =       {arXiv preprint arXiv:2504.07583},
  title =         {Do llms understand your translations? evaluating
                   paragraph-level mt with question answering},
  year =          {2025},
}

@article{papi2025hearing,
  author =        {Papi, Sara and Gilabert, Javier Garcia and
                   Hopton, Zachary and Zouhar, Vil{\'e}m and
                   Escolano, Carlos and G{\'a}llego, Gerard I and
                   Iranzo-S{\'a}nchez, Jorge and Kim, Ahrii and
                   Mach{\'a}{\v{c}}ek, Dominik and Schmidtova, Patricia and
                   others},
  journal =       {arXiv preprint arXiv:2512.16378},
  title =         {Hearing to Translate: The Effectiveness of Speech
                   Modality Integration into LLMs},
  year =          {2025},
}

@article{abouelenin2025phi,
  author =        {Abouelenin, Abdelrahman and Ashfaq, Atabak and
                   Atkinson, Adam and Awadalla, Hany and Bach, Nguyen and
                   Bao, Jianmin and Benhaim, Alon and Cai, Martin and
                   Chaudhary, Vishrav and Chen, Congcong and others},
  journal =       {arXiv preprint arXiv:2503.01743},
  title =         {Phi-4-mini technical report: Compact yet powerful
                   multimodal language models via mixture-of-loras},
  year =          {2025},
}

@article{liu2025voxtral,
  author =        {Liu, Alexander H and Ehrenberg, Andy and Lo, Andy and
                   Denoix, Cl{\'e}ment and Barreau, Corentin and
                   Lample, Guillaume and Delignon, Jean-Malo and
                   Chandu, Khyathi Raghavi and von Platen, Patrick and
                   Muddireddy, Pavankumar Reddy and others},
  journal =       {arXiv preprint arXiv:2507.13264},
  title =         {Voxtral},
  year =          {2025},
}

@inproceedings{lu2025developing,
  author =        {Lu, Ke-Han and Chen, Zhehuai and Fu, Szu-Wei and
                   Yang, Chao-Han Huck and Balam, Jagadeesh and
                   Ginsburg, Boris and Wang, Yu-Chiang Frank and
                   Lee, Hung-yi},
  booktitle =     {ICASSP 2025-2025 IEEE International Conference on
                   Acoustics, Speech and Signal Processing (ICASSP)},
  organization =  {IEEE},
  pages =         {1--5},
  title =         {Developing instruction-following speech language
                   model without speech instruction-tuning data},
  year =          {2025},
}

@inproceedings{radford2023robust,
  author =        {Radford, Alec and Kim, Jong Wook and Xu, Tao and
                   Brockman, Greg and McLeavey, Christine and
                   Sutskever, Ilya},
  booktitle =     {International conference on machine learning},
  organization =  {PMLR},
  pages =         {28492--28518},
  title =         {Robust speech recognition via large-scale weak
                   supervision},
  year =          {2023},
}

@article{rei2025tower+,
  author =        {Rei, Ricardo and Guerreiro, Nuno M and
                   Pombal, Jos{\'e} and Alves, Jo{\~a}o and
                   Teixeirinha, Pedro and Farajian, Amin and
                   Martins, Andr{\'e} FT},
  journal =       {arXiv preprint arXiv:2506.17080},
  title =         {Tower+: Bridging generality and translation
                   specialization in multilingual llms},
  year =          {2025},
}

@article{charness2012experimental,
  author =        {Charness, Gary and Gneezy, Uri and Kuhn, Michael A},
  journal =       {Journal of economic behavior \& organization},
  number =        {1},
  pages =         {1--8},
  publisher =     {Elsevier},
  title =         {Experimental methods: Between-subject and
                   within-subject design},
  volume =        {81},
  year =          {2012},
}

@article{brysbaert2019many,
  author =        {Brysbaert, Marc},
  journal =       {Journal of cognition},
  number =        {1},
  pages =         {16},
  title =         {How many participants do we have to include in
                   properly powered experiments? A tutorial of power
                   analysis with reference tables},
  volume =        {2},
  year =          {2019},
}

@inproceedings{lee-etal-2025-speech,
  address =       {Abu Dhabi, UAE},
  author =        {Lee, Beomseok and Gaido, Marco and Calapodescu, Ioan and
                   Besacier, Laurent and Negri, Matteo},
  booktitle =     {Proceedings of the 31st International Conference on
                   Computational Linguistics},
  editor =        {Rambow, Owen and Wanner, Leo and Apidianaki, Marianna and
                   Al-Khalifa, Hend and Eugenio, Barbara Di and
                   Schockaert, Steven},
  month =         jan,
  pages =         {6816--6826},
  publisher =     {Association for Computational Linguistics},
  title =         {Speech Foundation Models and Crowdsourcing for
                   Efficient, High-Quality Data Collection},
  year =          {2025},
  url =           {https://aclanthology.org/2025.coling-main.455/},
}

@article{bhattacherjee2001understanding,
  author =        {Bhattacherjee, Anol},
  journal =       {MIS quarterly},
  number =        {3},
  pages =         {351--370},
  publisher =     {Management Information Systems Research Center,
                   University of Minnesota},
  title =         {Understanding information systems continuance: An
                   expectation-confirmation model1},
  volume =        {25},
  year =          {2001},
}

@inproceedings{rei-etal-2020-comet,
  address =       {Online},
  author =        {Rei, Ricardo and Stewart, Craig and Farinha, Ana C and
                   Lavie, Alon},
  booktitle =     {Proceedings of the 2020 Conference on Empirical
                   Methods in Natural Language Processing (EMNLP)},
  editor =        {Webber, Bonnie and Cohn, Trevor and He, Yulan and
                   Liu, Yang},
  month =         nov,
  pages =         {2685--2702},
  publisher =     {Association for Computational Linguistics},
  title =         {{COMET}: A Neural Framework for {MT} Evaluation},
  year =          {2020},
  doi =           {10.18653/v1/2020.emnlp-main.213},
  url =           {https://aclanthology.org/2020.emnlp-main.213/},
}

@article{guerreiro-etal-2024-xcomet,
  address =       {Cambridge, MA},
  author =        {Guerreiro, Nuno M. and Rei, Ricardo and
                   Stigt, Daan van and Coheur, Luisa and Colombo, Pierre and
                   Martins, Andr{\'e} F. T.},
  journal =       {Transactions of the Association for Computational
                   Linguistics},
  pages =         {979--995},
  publisher =     {MIT Press},
  title =         {x{COMET}: Transparent Machine Translation Evaluation
                   through Fine-grained Error Detection},
  volume =        {12},
  year =          {2024},
  doi =           {10.1162/tacl_a_00683},
  url =           {https://aclanthology.org/2024.tacl-1.54/},
}

@inproceedings{juraska-etal-2024-metricx,
  address =       {Miami, Florida, USA},
  author =        {Juraska, Juraj and Deutsch, Daniel and
                   Finkelstein, Mara and Freitag, Markus},
  booktitle =     {Proceedings of the Ninth Conference on Machine
                   Translation},
  editor =        {Haddow, Barry and Kocmi, Tom and Koehn, Philipp and
                   Monz, Christof},
  month =         nov,
  pages =         {492--504},
  publisher =     {Association for Computational Linguistics},
  title =         {{M}etric{X}-24: The {G}oogle Submission to the {WMT}
                   2024 Metrics Shared Task},
  year =          {2024},
  doi =           {10.18653/v1/2024.wmt-1.35},
  url =           {https://aclanthology.org/2024.wmt-1.35/},
}

@inproceedings{fucci-etal-2025-different,
  address =       {Vienna, Austria},
  author =        {Fucci, Dennis and Gaido, Marco and Negri, Matteo and
                   Bentivogli, Luisa and Martins, Andre and
                   Attanasio, Giuseppe},
  booktitle =     {Proceedings of the 63rd Annual Meeting of the
                   Association for Computational Linguistics (Volume 2:
                   Short Papers)},
  editor =        {Che, Wanxiang and Nabende, Joyce and
                   Shutova, Ekaterina and Pilehvar, Mohammad Taher},
  month =         jul,
  pages =         {1005--1019},
  publisher =     {Association for Computational Linguistics},
  title =         {Different Speech Translation Models Encode and
                   Translate Speaker Gender Differently},
  year =          {2025},
  doi =           {10.18653/v1/2025.acl-short.78},
  isbn =          {979-8-89176-252-7},
  url =           {https://aclanthology.org/2025.acl-short.78/},
}

@inproceedings{kocmi-etal-2025-findings,
  address =       {Suzhou, China},
  author =        {Kocmi, Tom and Artemova, Ekaterina and
                   Avramidis, Eleftherios and Bawden, Rachel and
                   Bojar, Ond{\v{r}}ej and Dranch, Konstantin and
                   Dvorkovich, Anton and Dukanov, Sergey and
                   Fishel, Mark and Freitag, Markus and Gowda, Thamme and
                   Grundkiewicz, Roman and Haddow, Barry and
                   Karpinska, Marzena and Koehn, Philipp and
                   Lakougna, Howard and Lundin, Jessica and
                   Monz, Christof and Murray, Kenton and Nagata, Masaaki and
                   Perrella, Stefano and Proietti, Lorenzo and
                   Popel, Martin and Popovi{\'c}, Maja and Riley, Parker and
                   Shmatova, Mariya and Steingr{\'i}msson, Steinth{\'o}r and
                   Yankovskaya, Lisa and Zouhar, Vil{\'e}m},
  booktitle =     {Proceedings of the Tenth Conference on Machine
                   Translation},
  editor =        {Haddow, Barry and Kocmi, Tom and Koehn, Philipp and
                   Monz, Christof},
  month =         nov,
  pages =         {355--413},
  publisher =     {Association for Computational Linguistics},
  title =         {Findings of the {WMT}25 General Machine Translation
                   Shared Task: Time to Stop Evaluating on Easy Test
                   Sets},
  year =          {2025},
  doi =           {10.18653/v1/2025.wmt-1.22},
  isbn =          {979-8-89176-341-8},
  url =           {https://aclanthology.org/2025.wmt-1.22/},
}

@inproceedings{gaido-etal-2020-breeding,
  address =       {Barcelona, Spain (Online)},
  author =        {Gaido, Marco and Savoldi, Beatrice and
                   Bentivogli, Luisa and Negri, Matteo and
                   Turchi, Marco},
  booktitle =     {Proceedings of the 28th International Conference on
                   Computational Linguistics},
  editor =        {Scott, Donia and Bel, Nuria and Zong, Chengqing},
  month =         dec,
  pages =         {3951--3964},
  publisher =     {International Committee on Computational Linguistics},
  title =         {Breeding Gender-aware Direct Speech Translation
                   Systems},
  year =          {2020},
  doi =           {10.18653/v1/2020.coling-main.350},
  url =           {https://aclanthology.org/2020.coling-main.350/},
}

@inproceedings{savoldi-etal-2022-dynamics,
  address =       {Seattle, Washington},
  author =        {Savoldi, Beatrice and Gaido, Marco and
                   Bentivogli, Luisa and Negri, Matteo and
                   Turchi, Marco},
  booktitle =     {Proceedings of the 4th Workshop on Gender Bias in
                   Natural Language Processing (GeBNLP)},
  editor =        {Hardmeier, Christian and Basta, Christine and
                   Costa-juss{\`a}, Marta R. and Stanovsky, Gabriel and
                   Gonen, Hila},
  month =         jul,
  pages =         {94--111},
  publisher =     {Association for Computational Linguistics},
  title =         {On the Dynamics of Gender Learning in Speech
                   Translation},
  year =          {2022},
  doi =           {10.18653/v1/2022.gebnlp-1.12},
  url =           {https://aclanthology.org/2022.gebnlp-1.12/},
}

@inproceedings{savoldi-etal-2022-morphosyntactic,
  address =       {Dublin, Ireland},
  author =        {Savoldi, Beatrice and Gaido, Marco and
                   Bentivogli, Luisa and Negri, Matteo and
                   Turchi, Marco},
  booktitle =     {Proceedings of the 60th Annual Meeting of the
                   Association for Computational Linguistics (Volume 1:
                   Long Papers)},
  editor =        {Muresan, Smaranda and Nakov, Preslav and
                   Villavicencio, Aline},
  month =         may,
  pages =         {1807--1824},
  publisher =     {Association for Computational Linguistics},
  title =         {Under the Morphosyntactic Lens: A Multifaceted
                   Evaluation of Gender Bias in Speech Translation},
  year =          {2022},
  doi =           {10.18653/v1/2022.acl-long.127},
  url =           {https://aclanthology.org/2022.acl-long.127/},
}

@inproceedings{zhang2022mitigating,
  author =        {Zhang, Yuanyuan and Zhang, Yixuan and
                   Halpern, Bence Mark and Patel, Tanvina and
                   Scharenborg, Odette},
  booktitle =     {Interspeech},
  pages =         {3168--3172},
  title =         {Mitigating bias against non-native accents.},
  year =          {2022},
}

@inproceedings{harris-etal-2024-modeling,
  address =       {Miami, Florida, USA},
  author =        {Harris, Camille and Mgbahurike, Chijioke and
                   Kumar, Neha and Yang, Diyi},
  booktitle =     {Findings of the Association for Computational
                   Linguistics: EMNLP 2024},
  editor =        {Al-Onaizan, Yaser and Bansal, Mohit and
                   Chen, Yun-Nung},
  month =         nov,
  pages =         {15166--15184},
  publisher =     {Association for Computational Linguistics},
  title =         {Modeling Gender and Dialect Bias in Automatic Speech
                   Recognition},
  year =          {2024},
  doi =           {10.18653/v1/2024.findings-emnlp.890},
  url =           {https://aclanthology.org/2024.findings-emnlp.890/},
}

@inproceedings{attanasio-etal-2024-twists,
  address =       {Miami, Florida, USA},
  author =        {Attanasio, Giuseppe and Savoldi, Beatrice and
                   Fucci, Dennis and Hovy, Dirk},
  booktitle =     {Proceedings of the 2024 Conference on Empirical
                   Methods in Natural Language Processing},
  editor =        {Al-Onaizan, Yaser and Bansal, Mohit and
                   Chen, Yun-Nung},
  month =         nov,
  pages =         {21318--21340},
  publisher =     {Association for Computational Linguistics},
  title =         {Twists, Humps, and Pebbles: Multilingual Speech
                   Recognition Models Exhibit Gender Performance Gaps},
  year =          {2024},
  doi =           {10.18653/v1/2024.emnlp-main.1188},
  url =           {https://aclanthology.org/2024.emnlp-main.1188/},
}

@inproceedings{mehandru-etal-2023-physician,
  address =       {Singapore},
  author =        {Mehandru, Nikita and Agrawal, Sweta and Xiao, Yimin and
                   Gao, Ge and Khoong, Elaine and Carpuat, Marine and
                   Salehi, Niloufar},
  booktitle =     {Proceedings of the 2023 Conference on Empirical
                   Methods in Natural Language Processing},
  editor =        {Bouamor, Houda and Pino, Juan and Bali, Kalika},
  month =         dec,
  pages =         {11633--11647},
  publisher =     {Association for Computational Linguistics},
  title =         {Physician Detection of Clinical Harm in Machine
                   Translation: Quality Estimation Aids in Reliance and
                   Backtranslation Identifies Critical Errors},
  year =          {2023},
  doi =           {10.18653/v1/2023.emnlp-main.712},
  url =           {https://aclanthology.org/2023.emnlp-main.712/},
}

@article{valdez2025google,
  author =        {Valdez, Susana and Guerberof-Arenas, Ana},
  journal =       {Translation Spaces},
  number =        {2},
  pages =         {253--276},
  publisher =     {John Benjamins Publishing Company
                   Amsterdam/Philadelphia},
  title =         {“Google Translate is our best friend here” A
                   vignette-based interview study on machine translation
                   use for health communication},
  volume =        {14},
  year =          {2025},
}

@inproceedings{robertson2022understanding,
  author =        {Robertson, Samantha and D{\'\i}az, Mark},
  booktitle =     {Proceedings of the 2022 ACM conference on fairness,
                   accountability, and transparency},
  pages =         {2223--2238},
  title =         {Understanding and being understood: User strategies
                   for identifying and recovering from mistranslations
                   in machine translation-mediated chat},
  year =          {2022},
}

@inproceedings{xiao-etal-2025-toward,
  address =       {Suzhou, China},
  author =        {Xiao, Yimin and Zhang, Yongle and Ki, Dayeon and
                   Bao, Calvin and Martindale, Marianna J. and
                   Vaughn, Charlotte and Gao, Ge and Carpuat, Marine},
  booktitle =     {Proceedings of the 2025 Conference on Empirical
                   Methods in Natural Language Processing},
  editor =        {Christodoulopoulos, Christos and Chakraborty, Tanmoy and
                   Rose, Carolyn and Peng, Violet},
  month =         nov,
  pages =         {33997--34014},
  publisher =     {Association for Computational Linguistics},
  title =         {Toward Machine Translation Literacy: How Lay Users
                   Perceive and Rely on Imperfect Translations},
  year =          {2025},
  doi =           {10.18653/v1/2025.emnlp-main.1725},
  isbn =          {979-8-89176-332-6},
  url =           {https://aclanthology.org/2025.emnlp-main.1725/},
}

@inproceedings{ki-etal-2025-share,
  address =       {Suzhou, China},
  author =        {Ki, Dayeon and Duh, Kevin and Carpuat, Marine},
  booktitle =     {Proceedings of the 2025 Conference on Empirical
                   Methods in Natural Language Processing},
  editor =        {Christodoulopoulos, Christos and Chakraborty, Tanmoy and
                   Rose, Carolyn and Peng, Violet},
  month =         nov,
  pages =         {12069--12092},
  publisher =     {Association for Computational Linguistics},
  title =         {Should {I} Share this Translation? Evaluating Quality
                   Feedback for User Reliance on Machine Translation},
  year =          {2025},
  doi =           {10.18653/v1/2025.emnlp-main.606},
  isbn =          {979-8-89176-332-6},
  url =           {https://aclanthology.org/2025.emnlp-main.606/},
}

@article{mohammed2026far,
  author =        {Mohammed, Wafaa and Niculae, Vlad},
  journal =       {arXiv preprint arXiv:2606.16596},
  title =         {How Far Can Machine Translation Quality Take You?
                   Extrinsic Discourse Evaluation in Goal-Oriented
                   Setups},
  year =          {2026},
}

@inproceedings{liebling-etal-2022-opportunities,
  address =       {Seattle, United States},
  author =        {Liebling, Daniel and Heller, Katherine and
                   Robertson, Samantha and Deng, Wesley},
  booktitle =     {Findings of the Association for Computational
                   Linguistics: NAACL 2022},
  editor =        {Carpuat, Marine and de Marneffe, Marie-Catherine and
                   Meza Ruiz, Ivan Vladimir},
  month =         jul,
  pages =         {229--240},
  publisher =     {Association for Computational Linguistics},
  title =         {Opportunities for Human-centered Evaluation of
                   Machine Translation Systems},
  year =          {2022},
  doi =           {10.18653/v1/2022.findings-naacl.17},
  url =           {https://aclanthology.org/2022.findings-naacl.17/},
}

@inproceedings{ungless-etal-2025-amplifying,
  address =       {Vienna, Austria},
  author =        {Ungless, Eddie L. and Dev, Sunipa and
                   Bennett, Cynthia L. and Gulotta, Rebecca and
                   Bastings, Jasmijn and Denton, Remi},
  booktitle =     {Proceedings of the 63rd Annual Meeting of the
                   Association for Computational Linguistics (Volume 1:
                   Long Papers)},
  editor =        {Che, Wanxiang and Nabende, Joyce and
                   Shutova, Ekaterina and Pilehvar, Mohammad Taher},
  month =         jul,
  pages =         {20503--20535},
  publisher =     {Association for Computational Linguistics},
  title =         {Amplifying Trans and Nonbinary Voices: A
                   Community-Centred Harm Taxonomy for {LLM}s},
  year =          {2025},
  doi =           {10.18653/v1/2025.acl-long.1001},
  isbn =          {979-8-89176-251-0},
  url =           {https://aclanthology.org/2025.acl-long.1001/},
}

@inproceedings{belvin-etal-2005-transonics,
  address =       {Ann Arbor, Michigan},
  author =        {Belvin, Robert and Ettelaie, Emil and Gandhe, Sudeep and
                   Georgiou, Panayiotis and Knight, Kevin and
                   Marcu, Daniel and Millward, Scott and
                   Narayanan, Shrikanth and Neely, Howard and
                   Traum, David},
  booktitle =     {Proceedings of the {ACL} Interactive Poster and
                   Demonstration Sessions},
  editor =        {Nagata, Masaaki and Pedersen, Ted},
  month =         jun,
  pages =         {89--92},
  publisher =     {Association for Computational Linguistics},
  title =         {{T}ransonics: A Practical Speech-to-Speech Translator
                   for {E}nglish-{F}arsi Medical Dialogs},
  year =          {2005},
  doi =           {10.3115/1225753.1225776},
  url =           {https://aclanthology.org/P05-3023/},
}

@inproceedings{janakiram2020user,
  author =        {Janakiram, Antony Akash and Gerlach, Johanna and
                   Vuadens-Lehmann, Alyssa and Bouillon, Pierrette and
                   Spechbach, Herv{\'e}},
  booktitle =     {MIE},
  pages =         {1421--1422},
  title =         {User Satisfaction with a Speech-Enabled Translator in
                   Emergency Settings.},
  year =          {2020},
}

@article{fuchs2010differences,
  author =        {Fuchs, Susanne and Toda, Martine},
  journal =       {Turbulent sounds: An interdisciplinary guide},
  pages =         {281--302},
  publisher =     {Mouton de Gruyter Berlin, Germany},
  title =         {{Do differences in male versus female/s/reflect
                   biological or sociophonetic factors}},
  volume =        {21},
  year =          {2010},
}

@incollection{zimman2021gender,
  author =        {Zimman, Lal},
  booktitle =     {The Routledge handbook of language, gender, and
                   sexuality},
  pages =         {69--90},
  publisher =     {Routledge},
  title =         {Gender diversity and the voice},
  year =          {2021},
}

@article{zimman2020sociophonetics,
  author =        {Zimman, Lal},
  journal =       {The International Encyclopedia of Linguistic
                   Anthropology},
  pages =         {1--5},
  publisher =     {Wiley Online Library},
  title =         {Sociophonetics},
  year =          {2020},
}

@inproceedings{sanchez2024beyond,
  author =        {Sanchez, Ariadna and Ross, Alice and Markl, Nina},
  booktitle =     {2024 IEEE Spoken Language Technology Workshop (SLT)},
  organization =  {IEEE},
  pages =         {526--532},
  title =         {Beyond the binary: Limitations and possibilities of
                   gender-related speech technology research},
  year =          {2024},
}

@inproceedings{wolf-etal-2020-transformers,
  address =       {Online},
  author =        {Wolf, Thomas and Debut, Lysandre and Sanh, Victor and
                   Chaumond, Julien and Delangue, Clement and
                   Moi, Anthony and Cistac, Pierric and Rault, Tim and
                   Louf, Remi and Funtowicz, Morgan and Davison, Joe and
                   Shleifer, Sam and von Platen, Patrick and Ma, Clara and
                   Jernite, Yacine and Plu, Julien and Xu, Canwen and
                   Le Scao, Teven and Gugger, Sylvain and Drame, Mariama and
                   Lhoest, Quentin and Rush, Alexander},
  booktitle =     {Proceedings of the 2020 Conference on Empirical
                   Methods in Natural Language Processing: System
                   Demonstrations},
  editor =        {Liu, Qun and Schlangen, David},
  month =         oct,
  pages =         {38--45},
  publisher =     {Association for Computational Linguistics},
  title =         {Transformers: State-of-the-Art Natural Language
                   Processing},
  year =          {2020},
  doi =           {10.18653/v1/2020.emnlp-demos.6},
  url =           {https://aclanthology.org/2020.emnlp-demos.6/},
}

@article{jolly2018pymer4,
  author =        {Jolly, Eshin},
  journal =       {Journal of Open Source Software},
  number =        {31},
  pages =         {862},
  title =         {Pymer4: Connecting R and Python for linear mixed
                   modeling},
  volume =        {3},
  year =          {2018},
}
